\documentclass[journal,twoside,web]{ieeecolor}
\usepackage{tmi}
\usepackage{cite}
\usepackage{amsmath,amssymb,amsfonts}
\usepackage{algorithmic}
\usepackage{graphicx}
\usepackage{textcomp}
\usepackage{multicol}
\usepackage{multirow}
\usepackage{booktabs}
\usepackage[table]{xcolor}
\usepackage{makecell}
\usepackage{tabularx}
\usepackage{mwe}
\usepackage{soul}

\def\BibTeX{{\rm B\kern-.05em{\sc i\kern-.025em b}\kern-.08em
    T\kern-.1667em\lower.7ex\hbox{E}\kern-.125emX}}
\begin{document}
\title{EviATTA: Evidential Active Test-Time Adaptation for Medical Segment Anything Models}
\author{Jiayi Chen, Yasmeen George, Winston Chong, and Jianfei Cai, \IEEEmembership{Fellow, IEEE}
\thanks{This work was supported by the Medical Research Future Fund (MRFF) under Grant NCRI000074. (Corresponding author: Jiayi Chen)}
\thanks{Jiayi Chen, Yasmeen George,  and Jianfei Cai are with Department of Data Science \& AI, Faculty of Information Technology, Monash University, VIC 3800, Australia (e-mail: \{jiayi.chen, yasmeen.george, jianfei.cai\}@monash.edu)}
\thanks{Winston Chong is with Alfred Health, VIC 3004, Australia (e-mail: winston.chong@monash.edu)}
}

\maketitle

\begin{abstract}
Deploying foundational medical Segment Anything Models (SAMs) via test-time adaptation (TTA) is challenging under large distribution shifts, where test-time supervision is often unreliable. While active test-time adaptation (ATTA) introduces limited expert feedback to improve reliability, existing ATTA methods still suffer from unreliable uncertainty estimation and inefficient utilization of sparse annotations. To address these issues, we propose Evidential Active Test-Time Adaptation (EviATTA), which is, to our knowledge, the first ATTA framework tailored for medical SAMs. Specifically, we adopt the Dirichlet-based Evidential Modeling to decompose overall predictive uncertainty into distribution uncertainty and data uncertainty. Building on this decomposition, we design a Hierarchical Evidential Sampling strategy, where image-wise distribution uncertainty is used to select informative shifted samples, while distance-aware data uncertainty guides sparse pixel annotations to resolve data ambiguities. We further introduce Dual Consistency Regularization, which enforces progressive prompt consistency on sparsely labeled samples to better exploit sparse supervision and applies variational feature consistency on unlabeled samples to stabilize adaptation. Extensive experiments on six medical image segmentation datasets demonstrate that EviATTA consistently improves adaptation reliability with minimal expert feedback under both batch-wise and instance-wise test-time adaptation settings.

\end{abstract}


\section{Introduction}
\begin{figure*}[t]
    \centering
    \includegraphics[width=0.95\linewidth]{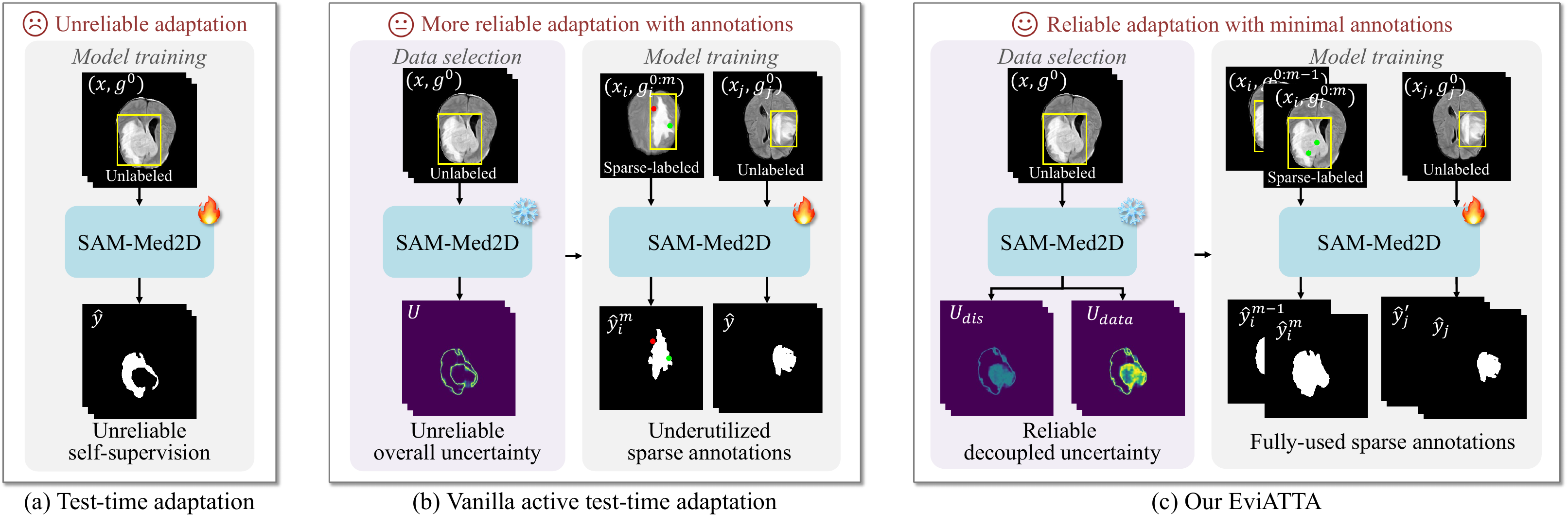}
    \caption{Comparison of TTA, vanilla ATTA, and EviATTA. (a) TTA suffers from unreliable adaptation due to unreliable self-supervision. (b) Vanilla ATTA partially improves adaptation but still struggles with unreliable uncertainty estimation and inefficient utilization of sparse annotations. (c) EviATTA employs decoupled uncertainty for more reliable uncertainty estimation and leverages dual consistency regularization to exploit sparse annotations.}
    \label{fig:open}
\end{figure*}

Recently, medical segment anything models (SAM), such as SAM-Med2D~\cite{cheng2023sammed2d}, MedSAM~\cite{ma2024segment}, and their variants~\cite{wu2025medical,zhu2024medical}, have emerged as strong foundation models for medical image segmentation. They provide flexible prompt-based interaction and show strong zero-shot generalization across organs and lesions~\cite{huang2024segment}. However, their performance often degrades in real clinical deployment due to distribution shifts arising from variations in imaging modalities, acquisition protocols~\cite{ma2025tsar}, patient populations \cite{xia2024enhancing}, or anatomical structures~\cite{chen2025test,fu2024cosam,ma2023federated,huang2025fly}.

Test-time adaptation (TTA) is a promising way to mitigate distribution shifts by adapting pretrained models during inference without labels~\cite{liang2025comprehensive}. As shown in Fig.~\ref{fig:open}(a), existing TTA methods mainly rely on self-training signals, such as entropy minimization~\cite{wang2020tent,zhangcome,chen2025gradient} and consistency regularization~\cite{wang2022continual,niuself,chen2025gradient,wu2025sam}. Under large distribution shifts, these signals are often noisy and miscalibrated, leading to error accumulation and even negative adaptation~\cite{wang2025effortless}. Prior studies attempt to mitigate this issue with reliable sample selection~\cite{niutowards}, pseudo-label refinement~\cite{chen2022contrastive,huselective}, and anti-forgetting mechanisms~\cite{niu2022efficient}. However, these methods remain limited to model-generated supervision, rendering the adaptation intrinsically unstable without reliable external guidance.

To improve adaptation reliability, active test-time adaptation (ATTA)~\cite{gui2024active} integrates active learning~\cite{wang2024comprehensive} into TTA. As illustrated in Fig.~\ref{fig:open}(b), ATTA selects a limited number of informative test samples or pixels for annotation to reduce error accumulation. Most ATTA methods focus on image classification, where image-wise informativeness is evaluated using feature diversity~\cite{gui2024active,liexploring} and predictive uncertainty, typically quantified via entropy~\cite{gui2024active,leetest,tanexposing}, softmax confidence~\cite{liexploring}, conformal score~\cite{shi2025annotation}, or perturbation consistency~\cite{wang2025effortless}. While a few efforts extend ATTA to segmentation~\cite{yuan2023few,islam2025odes}, they are built on non-promptable architectures, leaving ATTA for promptable foundation models underexplored.

Applying ATTA to medical SAMs raises two key challenges:
\textbf{(1) Unreliable uncertainty estimation.} 
Most ATTA methods estimate uncertainty using predictive entropy or softmax confidence. Under pronounced distribution shifts, model predictions often become overconfident and miscalibrated~\cite{watson2023explaining,chen2024think}, which makes these uncertainty proxies less reliable. As a result, active sample selection becomes less reliable, and critical samples that require correction are often overlooked.
\textbf{(2) Inefficient utilization of sparse annotations.} 
Classification-oriented methods rely on image-wise annotations and therefore lack the spatial granularity needed by segmentation. Segmentation-oriented methods, in contrast, are usually developed on non-promptable architectures and treat sparse pixel labels as isolated supervision points~\cite{yuan2023few}. Consequently, the former cannot precisely localize where to query annotations, and the latter cannot fully convert sparse annotations into dense, prompt-driven guidance in SAM.

To address these challenges, we propose \textbf{Evi}dential \textbf{A}ctive \textbf{T}est-\textbf{T}ime \textbf{A}daptation (EviATTA), which is, to our knowledge, the first ATTA framework tailored for medical SAMs. As illustrated in Fig.~\ref{fig:open}(c), we first introduce Dirichlet-based Evidential Modeling to decompose predictive uncertainty into distribution and data components~\cite{sensoy2018evidential,chen2024fedevi}. Based on this decomposition, we design a Hierarchical Evidential Sampling strategy that utilizes image-wise distribution uncertainty to select highly shifted samples and employ distance-aware data uncertainty to guide sparse annotations. We further propose Dual Consistency Regularization to improve sparse label utilization and adaptation stability. For sparsely labeled samples, we treat annotated pixels as spatial point prompts and enforce progressive consistency between weak and strong prompts, thereby propagating sparse cues into dense supervision. Meanwhile, we leverage variational feature consistency to regularize the adaptation dynamics.

Our main contributions are summarized as follows:
\begin{itemize}
    \item We propose EviATTA, the first active test-time adaptation framework tailored for medical SAMs, which improves adaptation reliability with minimal annotation cost.
    \item We introduce Dirichlet-based evidential modeling to decompose predictive uncertainty into distribution and data components, and propose a hierarchical evidential sampling strategy for reliable sample and pixel selection.
    \item We devise a dual consistency regularization scheme that treats sparse annotations as point prompts, improves sparse supervision utilization, and stabilizes adaptation.
    \item Extensive experiments on six medical image segmentation datasets demonstrate that EviATTA consistently outperforms state-of-the-art methods in both batch-wise and instance-wise test-time adaptation settings.
\end{itemize}

\section{Related Work}
\subsection{Test-Time Adaptation} 
Test-time adaptation is a paradigm to mitigate distribution shifts by adapting pre-trained models using unlabeled test samples. Existing TTA methods can be broadly categorized into two groups: entropy minimization~\cite{wang2020tent,niutowards,zhangcome} and consistency-based methods~\cite{wang2022continual,niuself,chen2025gradient}. \textbf{Entropy minimization methods} adapt pretrained models by reducing the entropy of model predictions on test data. For instance, TENT~\cite{wang2020tent} proposes a test-time entropy minimization strategy to update the affine parameters of batch normalization (BN) layers. SAR~\cite{niutowards} filters noisy samples with large gradients and employs sharpness-aware minimization to improve the stability of entropy minimization. Furthermore, COME~\cite{zhangcome} leverages a Dirichlet prior distribution to model subjective opinions and minimizes the entropy of opinions to boost reliability. 
\textbf{Consistency-based methods} exploit self-consistency to provide supervisory signals. CoTTA~\cite{wang2022continual} adopts the teacher-student framework to enforce prediction consistency between a test sample and its augmentations. SPA~\cite{niuself} introduces geometry-preserving augmentations, including low-frequency amplitude masking and high-frequency noise injection, to optimize prediction consistency across deteriorated views. Additionally, GraTa~\cite{chen2025gradient} achieves effective adaptation by aligning the gradients of the empirical loss and pseudo loss.

However, both entropy-minimization and consistency-based methods fundamentally rely on self-generated signals (\textit{e.g.}, predictive entropy or pseudo labels), which tend to be miscalibrated and noisy under distribution shifts~\cite{xiedirichlet}. Consequently, the adaptation process is driven by unreliable or misleading signals, which leads to error accumulation and unreliable adaptation~\cite{watson2023explaining,chen2024think}. Although several approaches incorporate reliable sample selection~\cite{niutowards}, pseudo label refinement~\cite{huselective}, and anti-forgetting mechanisms~\cite{niu2022efficient}, existing TTA methods struggle with error accumulation due to the lack of reliable supervision. This motivates the integration of minimal expert guidance to improve adaptation reliability~\cite{gui2024active,hu2024towards}.

\begin{figure*}[t]
    \centering
    \includegraphics[width=\linewidth]{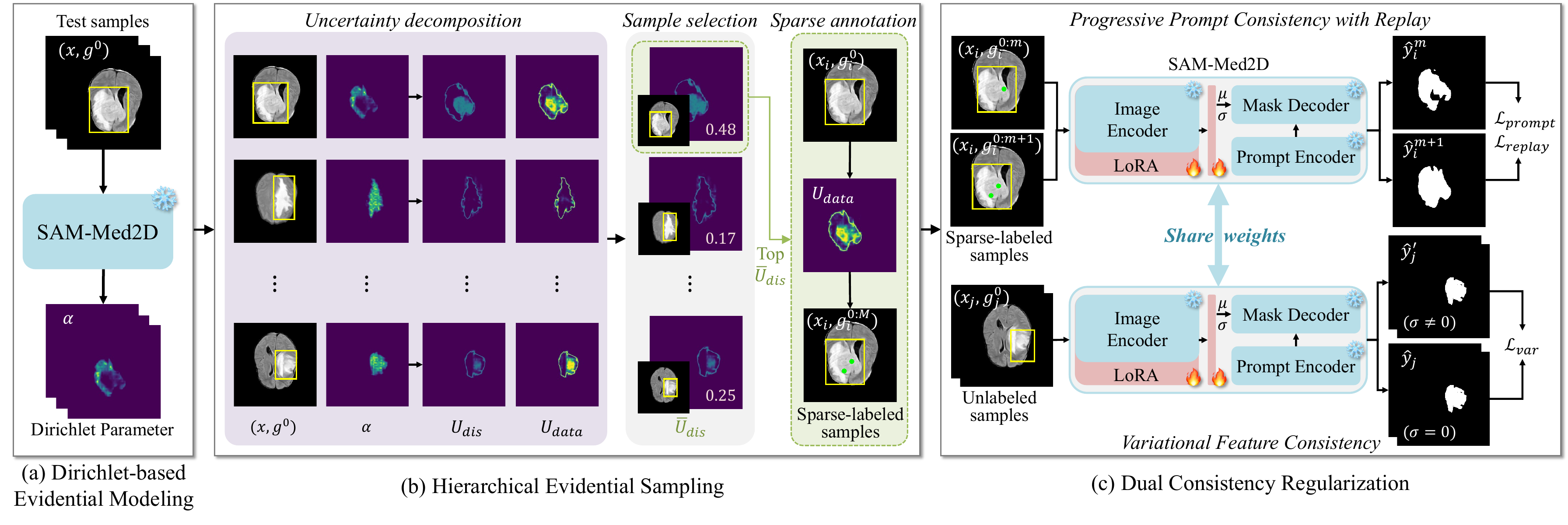}
    \caption{Framework of EviATTA. It comprises three components: (a) Dirichlet-based Evidential Modeling, which formulates categorical prediction as following a Dirichlet distribution; (b) Hierarchical Evidential Sampling, which decomposes overall uncertainty into distribution and data uncertainties, selects samples via image-wise distribution uncertainty, and annotates pixels using pixel-wise data uncertainty; (c) Dual Consistency Regularization, which enforces progressive prompt consistency to leverage sparse annotations and variational feature consistency to stabilize adaptation.}
    \label{fig:framework}
\end{figure*}
\subsection{Active Test-Time Adaptation}
Active test-time adaptation (ATTA) incorporates active learning into test-time adaptation, where a limited number of informative samples or pixels are annotated to alleviate error accumulation and facilitate more reliable adaptation under distribution shifts~\cite{gui2024active}. Unlike active learning that queries samples during offline training~\cite{wang2024comprehensive,chen2025active}, active test-time adaptation selects samples during online inference under strict annotation budgets, making the problem more challenging and still underexplored. 
Recent efforts have investigated ATTA for general classification tasks. For instance, SimATTA~\cite{gui2024active} employs incremental clustering on highly uncertain test samples for real-time sample selection. HiATTA~\cite{liexploring} proposes a k-margin strategy to annotate low-confident and diverse samples and subsequently utilizes the labeled samples for hyper-parameter selection. EATTA~\cite{wang2025effortless} identifies valuable samples via feature perturbation and dynamically balances optimization between labeled and unlabeled data. BiTTA~\cite{leetest} estimates data uncertainty via predictive entropy and requests binary feedback on uncertain samples to verify model predictions. In contrast, only a few attempts extend ATTA to segmentation tasks~\cite{yuan2023few,islam2025odes}. Both ATASeg~\cite{yuan2023few} and ODES~\cite{islam2025odes} estimate pixel-wise informativeness based on predictive entropy and region impurity to query sparse pixel annotations. ODES further introduces a sample selection strategy using BN statistics to prioritize samples with large distribution shifts, extending ATASeg to batch-wise adaptation. However, both methods are inherently built upon conventional non-promptable architectures, leaving active test-time adaptation for promptable foundation models unexplored.

Despite their effectiveness, existing ATTA approaches still suffer from unreliable uncertainty estimation and inefficient utilization of sparse annotations. In this paper, we propose a hierarchical evidential sampling strategy for more reliable uncertainty estimation and a dual consistency regularization scheme to exploit sparse annotation and stabilize adaptation.

\section{Method}
\subsection{Overview}
The overview of EviATTA is illustrated in Fig.~\ref{fig:framework}. Given a pre-trained medical SAM, EviATTA aims to adapt it to a target domain in an online test-time setting with minimal annotation cost. The medical SAM $\Theta=\{E_I, E_P, D\}$ comprises an image encoder $E_I(\cdot)$, a prompt encoder $E_P(\cdot)$, and a mask decoder $D(\cdot)$. During active test-time adaptation, unlabeled test samples are processed sequentially in mini-batches $\mathcal{B}=\{(\boldsymbol{x}_i,\boldsymbol{g}_i^0)\}_{i=1}^N$, where $N$ is the batch size, $\boldsymbol{x}_i\in\mathbb{R}^{H \times W\times 3}$ is the input image and $\boldsymbol{g}_i^0$ denotes the initial prompt. For each sample $(\boldsymbol{x}_i,\boldsymbol{g}_i^0)$, SAM produces an initial prediction $\hat{\boldsymbol{y}}_i$. According to Dirichlet-based evidential modeling, we decompose the overall uncertainty into distribution and data uncertainties, laying a foundation for detailed uncertainty estimation. To ensure a more comprehensive and reliable uncertainty assessment, we then perform hierarchical evidential sampling to select top-$K$ shifted samples via image-wise distribution uncertainty and annotate $M$ ambiguous pixels per selected sample via distance-aware data uncertainty. The annotated pixels are explicitly treated as point prompts, forming a sparsely labeled set $\mathcal{B}_L=\{(\boldsymbol{x}_i, \boldsymbol{g}_i^{0:M})\}_{i=1}^K$, while the remaining samples form the unlabeled set $\mathcal{B}_U={\mathcal{B}\setminus \mathcal{B}_{L}}=\{(\boldsymbol{x}_j, \boldsymbol{g}_j^{0})\}_{j=1}^{N-K}$. We further introduce dual consistency regularization to exploit sparse annotations and stabilize adaptation. Specifically, sparsely labeled samples $\mathcal{B}_L$ are adapted via progressive prompt consistency with replay, while unlabeled samples $\mathcal{B}_U$ are regularized through variational feature consistency.

\subsection{Dirichlet-based Evidential Modeling}
\label{sec:dem}
As shown in Fig.~\ref{fig:framework}(a), we introduce Dirichlet-based evidential modeling for medical SAMs to establish a theoretical foundation for detailed uncertainty decomposition.
Formally, given an input image $\boldsymbol{x}$ from the target domain and its prompt $\boldsymbol{g}$, the medical SAM $\Theta$ processes them through an image encoder $E_I$ and a prompt encoder $E_P$, yielding the image embedding $E_I(\boldsymbol{x})$ and prompt embedding $E_P(\boldsymbol{g})$. Then, the mask decoder $D$ projects these embeddings to spatial logits $\hat{\boldsymbol{y}}\in \mathbb{R}^{H\times W\times C}$, where $\hat{\boldsymbol{y}}=\Theta(\boldsymbol{x},\boldsymbol{g})=D(E_I(\boldsymbol{x}),E_P(\boldsymbol{g}))$. The conventional SAM utilizes softmax or sigmoid operators to map the logits into deterministic categorical probabilities $\boldsymbol{\rho}$. However, relying on such point estimates makes the model prone to overconfidence during test-time adaptation with distribution shifts~\cite{xiedirichlet}. To address with this issue, evidential modeling formulates the categorical prediction $\boldsymbol{\rho}$ as a random variable following a Dirichlet distribution. Given the input data $(\boldsymbol{x},\boldsymbol{g})$ and SAM $\Theta$, the probability density function of $\boldsymbol{\rho}$~\cite{sensoy2018evidential} is formulated as:
\begin{equation}
    p(\boldsymbol{\rho}|\boldsymbol{x},\boldsymbol{g},\Theta)=\left\{
    \begin{aligned}
         \frac{\Gamma(\sum_{c=1}^C\alpha_c)}{\prod_{c=1}^C\Gamma(\alpha_c)}\prod_{c=1}^C\rho_c^{\alpha_c-1}&, (\boldsymbol{\rho}\in \Delta^C) \\
         0\qquad\qquad&, (\text{otherwise}) 
    \end{aligned}
    \right.
\end{equation}
where $C$ is the number of classes, $\Gamma(\cdot)$ denotes the Gamma function, and $\boldsymbol{\alpha}\in \mathbb{R}^{H\times W\times C}$ is the Dirichlet parameter for the input data $(\boldsymbol{x},\boldsymbol{g})$. Based on subjective logic~\cite{josang2016subjective}, the parameter $\boldsymbol{\alpha}$ is derived as $\boldsymbol{\alpha}=\boldsymbol{e}+1=\mathcal{A}(\hat{\boldsymbol{y}})+1$, where $\mathcal{A}(\cdot)$ is the non-negative activation function. In this study, we implement $\mathcal{A}(\cdot)$ as the exponential function $\text{exp}(\cdot)$. 

Accordingly, the posterior probability for class $c$~\cite{xiedirichlet} is
\begin{equation}
    P(\boldsymbol{y}=c|\boldsymbol{x},\boldsymbol{g},\Theta)=\int p(\boldsymbol{y}=c|\boldsymbol{\rho})\cdot  p(\boldsymbol{\rho}|\boldsymbol{x},\boldsymbol{g},\Theta) \text{d}\boldsymbol{\rho}=\frac{\alpha_c}{S},
\end{equation}
where $S=\sum_{c=1}^C\alpha_c$ is the Dirichlet strength.

\subsection{Hierarchical Evidential Sampling}
\label{sec:hes}
As shown in Fig.~\ref{fig:framework}(b), we then leverage the evidential framework to decompose the overall predictive uncertainty into distribution and data components to capture distribution shifts and inherent complexity, respectively. Subsequently, we propose a hierarchical evidential sampling strategy that simultaneously considers image-wise distribution shifts and pixel-wise data ambiguity for more reliable uncertainty estimation.

\subsubsection{Uncertainty Decomposition}
Given the input data $(\boldsymbol{x},\boldsymbol{g})$ and SAM $\Theta$, the overall predictive uncertainty  $\boldsymbol{U}\in\mathbb{R}^{H\times W}$ is quantified as the entropy of posterior probability~\cite{malinin2018predictive}.
\begin{equation}
\begin{aligned}
    \boldsymbol{U}&=\mathcal{H}[P(\boldsymbol{y}|\boldsymbol{x}, \boldsymbol{g},\Theta)]=-\sum_{c=1}^C\overline{\rho}_c\log{\overline{\rho}_c}.
\end{aligned}
\end{equation}
From an information-theoretic perspective, the overall uncertainty can be decomposed into distribution uncertainty and data uncertainty to measure the distribution shifts and inherent ambiguity, respectively~\cite{malinin2018predictive,chen2024think}. Specifically, the data uncertainty $\boldsymbol{U}_{data}\in\mathbb{R}^{H\times W}$ is estimated as the expected entropy of all possible predictions:
\begin{equation}
    \boldsymbol{U}_{data}=\mathbb{E}[\mathcal{H}[P(\boldsymbol{y}|\boldsymbol{\rho})]]=\sum_{c=1}^C \overline{\rho}_{c}(\psi(S+1)-\psi(\alpha_c+1)),
\end{equation}
where $\mathcal{H}$ denotes the entropy and $\psi(\cdot)$ is the Digamma function. The distribution uncertainty $\boldsymbol{U}_{dis}\in\mathbb{R}^{H\times W}$ is formulated as the mutual information between categorical label $\boldsymbol{y}$ and probability $\boldsymbol{\rho}$~\cite{xiedirichlet,malinin2018predictive}, which is measured as the difference between the overall uncertainty and data uncertainty:
\begin{equation}
\small
    \boldsymbol{U}_{dis}=\boldsymbol{U}-\boldsymbol{U}_{data}=\sum_{c=1}^C \overline{\rho}_{c}\big(\psi(\alpha_c+1)-\psi(S+1)-\log{\overline{\rho}_c}\big).
\end{equation}

\subsubsection{Sampling Strategy}
To ensure reliable uncertainty assessment, it is crucial to consider both the distribution shifts from the pre-trained model and the data ambiguity of test samples. Therefore, we propose a hierarchical sampling strategy that first leverages image-wise distribution uncertainty to identify samples with notable distribution shifts, and then utilizes pixel-wise data uncertainty to pinpoint ambiguous regions for sparse annotation. Specifically, we quantify the image-wise distribution uncertainty $\overline{U}_{dis}$ by spatially aggregating the dense distribution uncertainty map $\boldsymbol{U}_{dis}$, which is defined as:
\begin{equation}
    \overline{U}_{dis}=\frac{1}{H\times W}\sum_{h=1}^H\sum_{w=1}^W {\boldsymbol{U}_{dis}^{h,w}}.
\end{equation}
Accordingly, the top-$K$ samples with the highest $\overline{U}_{dis}$ are selected for further annotation.

To minimize annotation costs, we leverage the promptable nature of SAM by adopting sparse annotations as auxiliary point prompts $\boldsymbol{g}$. To identify the most informative pixels, we jointly consider individual data uncertainty and spatial coverage. Specifically, we binarize the data uncertainty map $\boldsymbol{U}_{data}$ via Otsu's adaptive thresholding $\tau$~\cite{otsu1979threshold}, explicitly partitioning the map into uncertain and confident regions. We then apply the distance transform $\mathrm{Dist}(\cdot)$ to compute a spatial distance map $\boldsymbol{d}\in\mathbb{R}^{H\times W}$, which encodes the shortest distance from each pixel within the uncertain region ($\boldsymbol{U}_{data}{>}\tau$) to its nearest confident pixel. The distance map $\boldsymbol{d}$ is defined as:
\begin{equation}
    \boldsymbol{d}=\mathrm{Dist}(\mathbb{I}(\boldsymbol{U}_{data}>\tau)).
\end{equation}
Subsequently, we derive a distance-aware data uncertainty map via element-wise multiplication $\boldsymbol{d} \odot \boldsymbol{U}_{data}$. The coordinates $(h^*, w^*)$ of the next annotated pixel correspond to the spatial location of the maximum value within this reweighted map:
\begin{equation}
    (h^*,w^*) = \mathop{\arg\max}_{h,w} \; d^{h,w} \cdot U_{data}^{h,w}.
    \label{eq:selection}
\end{equation}

At the $m$-th iteration, the annotated pixel $(h^*,w^*)$ is incorporated into the prompt set, updating it to $\boldsymbol{g}^{0:m}$. Consequently, the segmentation mask and data uncertainty map are dynamically re-estimated. Guided by the updated data uncertainty, the $(m+1)$-th pixel is then queried via Eq.~\ref{eq:selection}. This iterative procedure continues until a budget of $M$ pixels is reached, inherently ensuring both high informativeness and broad spatial coverage for effective supervision.

\subsection{Dual Consistency Regularization}
\label{sec:dcr}
Given the limited sparse annotations, we design a dual consistency regularization scheme (see Fig.~\ref{fig:framework}(c)). It enforces progressive prompt consistency to extensively exploit sparse annotations and applies variational feature consistency to stabilize model adaptation.

\subsubsection{Progressive Prompt Consistency with Replay}
To extensively exploit sparse annotations, we enforce consistency across predictions derived from progressively refined prompts. For an input image $\boldsymbol{x}_i$, let $\boldsymbol{g}_i^0$ denote the initial box prompt, and $\boldsymbol{g}_i^{0:m}$ denote the prompt augmented with the first $m$ annotated points. 
This formulation yields a sequence of weak-strong prompt pairs $(\boldsymbol{g}_i^{0:m},\, \boldsymbol{g}_i^{0:m+1})$. By treating the prediction from the stronger prompt $\hat{\boldsymbol{y}}_i^{m+1}=\Theta(\boldsymbol{x}_i,\boldsymbol{g}_i^{0:m+1})$ as a dynamic target, we distill its knowledge into the weaker prediction $\hat{\boldsymbol{y}}_i^{m}=\Theta(\boldsymbol{x}_i,\boldsymbol{g_i}^{0:m})$. The progressive prompt consistency is defined as:
\begin{equation}
\small
\mathcal{L}_{prompt} = \sum_{m=0}^{M-1} \mathbb{I}_{conf}(\hat{\boldsymbol{y}}_i^{m+1},\hat{\boldsymbol{y}}_i^{m}) \cdot KL\big(\boldsymbol{sg}(\hat{\boldsymbol{y}}_i^{m+1}) \parallel \hat{\boldsymbol{y}}_i^m\big),
\end{equation}
\begin{equation}
\mathbb{I}_{conf}(\hat{\boldsymbol{y}}_i^{m+1},\hat{\boldsymbol{y}}_i^{m}) = \mathbf{1}\big[\max_c(\hat{\boldsymbol{y}}_{i,c}^{m+1}) > \max_c(\hat{\boldsymbol{y}}_{i,c}^m)\big],
\end{equation}
where $\mathbb{I}_{conf}(\hat{\boldsymbol{y}}_i^{m+1},\hat{\boldsymbol{y}}_i^{m})$ is an indicator to identify pixels with higher confidence in $\hat{\boldsymbol{y}}_i^{m+1}$ than in $\hat{\boldsymbol{y}}_i^{m}$. $\boldsymbol{sg}(\cdot)$ denotes the stop-gradient operation. To consolidate the knowledge gained from minimal expert feedback, we maintain a replay buffer that stores sparsely labeled samples along with their historical predictions $\hat{\boldsymbol{y}}_r$, enforcing consistency with the current predictions $\hat{\boldsymbol{y}}_r'$. The replay loss is defined as:
\begin{equation}
    \mathcal{L}_{replay}=\mathbb{I}_{conf}(\hat{\boldsymbol{y}}_r,\hat{\boldsymbol{y}}_r')\cdot KL(\boldsymbol{sg}(\hat{\boldsymbol{y}}_r)\parallel\hat{\boldsymbol{y}}_r'),
\end{equation}
where $\mathbb{I}_{conf}(\hat{\boldsymbol{y}}_r,\hat{\boldsymbol{y}}_r')$ is an indicator to filter reliable pixels.

\subsubsection{Variational Feature Consistency}
For unlabeled samples, we introduce variational feature consistency to stabilize adaptation. Rather than a deterministic embedding, we model the image feature as a latent Gaussian distribution $\mathcal{N}(\mu(\boldsymbol{x}_j), \sigma^2(\boldsymbol{x}_j))$. Here, the mean $\mu(\boldsymbol{x}_j)= E_I(\boldsymbol{x}_j)$ is the encoder output, and the standard deviation $\sigma(\boldsymbol{x}_j)$ is estimated via an auxiliary convolutional head~\cite{ma2024vnas}. Using the standard reparameterization trick, we sample a perturbed feature $\boldsymbol{z}_j$ as:
\begin{equation}
\boldsymbol{z}_j = E_I(\boldsymbol{x}_j) + \boldsymbol{\epsilon} \odot \sigma(\boldsymbol{x}_j), \quad \boldsymbol{\epsilon} \sim \mathcal{N}(\mathbf{0}, \mathbf{I}),
\end{equation}
where $\odot$ denotes element-wise multiplication. The mask decoder $D$ then generates a variance-augmented prediction based on the sampled feature and the initial prompt $\boldsymbol{g}_j^0$:
\begin{equation}
\hat{\boldsymbol{y}}_j' = D(\boldsymbol{z}_j, E_P(\boldsymbol{g}_j^0)).
\end{equation}
We subsequently enforce consistency between this perturbed output $\hat{\boldsymbol{y}}_j'$ and the original deterministic prediction $\hat{\boldsymbol{y}}_j=D(E_I(\boldsymbol{x}_j),E_P(\boldsymbol{g}_j))$. Thus, the variational feature consistency loss is defined as:
\begin{equation}
    \mathcal{L}_{var} = \mathbb{I}_{conf}(\hat{\boldsymbol{y}}_j,\hat{\boldsymbol{y}}_j')\cdot KL(\boldsymbol{sg}(\hat{\boldsymbol{y}}_j)\parallel\hat{\boldsymbol{y}}_j'),
\end{equation}
where $\mathbb{I}_{conf}(\hat{\boldsymbol{y}}_j,\hat{\boldsymbol{y}}_j')$ is the indicator to select reliable pixels.

\subsection{Parameter-Efficient Training Scheme}
During active test-time adaptation, we introduce LoRA modules to the output projection weight $\boldsymbol{W}_O$ in each self-attention module of the image encoder, while keeping the prompt encoder and mask decoder frozen. The adapted output projection is formulated as
\begin{equation}
    \boldsymbol{W}=\boldsymbol{W}_O+\Delta \boldsymbol{W}=\boldsymbol{W}_O+\boldsymbol{B}\boldsymbol{A},
\end{equation}
where $\boldsymbol{A}\in \mathbb{R}^{r\times d}$ and $\boldsymbol{B}\in\mathbb{R}^{d\times r}$ are low-rank matrices.

The overall loss function integrates the progressive prompt consistency (including replay) for sparsely labeled samples and the variational feature consistency for unlabeled samples, formulated as follows:
\begin{equation}
    \mathcal{L}=\mathcal{L}_{prompt}+\mathcal{L}_{replay}+\mathcal{L}_{var}.
\end{equation}

\section{Experiment}
\begin{table*}[htbp]
    \centering
    \caption{Performance comparison with SOTA TTA and ATTA methods under batch-wise setting. Best results are highlighted as \colorbox{red!20}{First}, \colorbox{orange!20}{Second}, and \colorbox{yellow!20}{Third}. $10\%$ and $20\%$ indicate selecting 10\% and 20\% samples per batch for sparse annotation.}
    \setlength{\tabcolsep}{2mm}
    \begin{tabularx}{\linewidth}{clcccccccccccc}
        \bottomrule
         \multicolumn{1}{c}{\multirow{2}{*}{Type}} & \multicolumn{1}{c}{\multirow{2}{*}{Method}} & \multicolumn{4}{c}{BraTS-SSA} & \multicolumn{4}{c}{BraTS-PED} & \multicolumn{4}{c}{FHPsAOP}\\
        \cmidrule(lr){3-6}\cmidrule(lr){7-10}\cmidrule(lr){11-14}
        & & Dice$\uparrow$ & Jaccard$\uparrow$ & ASD$\downarrow$ & HD95$\downarrow$ & Dice$\uparrow$ & Jaccard$\uparrow$ & ASD$\downarrow$ & HD95$\downarrow$ & Dice$\uparrow$ & Jaccard$\uparrow$ & ASD$\downarrow$ & HD95$\downarrow$\\
        \hline
        Zero-shot & SAM-Med2D~\cite{cheng2023sammed2d} & \cellcolor{gray!20}82.34 & \cellcolor{gray!20}71.64 & \cellcolor{gray!20}2.73 & \cellcolor{gray!20}13.10 & \cellcolor{gray!20}84.84 & \cellcolor{gray!20}75.32 & \cellcolor{gray!20}2.12 & \cellcolor{gray!20}7.77 & \cellcolor{gray!20}73.87 & \cellcolor{gray!20}60.84 & \cellcolor{gray!20}11.16 & \cellcolor{gray!20}29.58\\
        
        \hline
        \multirow{6}{*}{TTA} & TENT~\cite{wang2020tent} & 81.67 & 70.81 & 2.75 & 13.82 & 86.30 & 77.39 & 1.96 & 7.55 & 67.66 & 52.36 & 12.76 & 34.24\\
        & CoTTA~\cite{wang2022continual} & 84.09 & 73.90 & 2.70 & 13.01 & 85.87 & 76.53 & 2.16 & 7.74 & 74.19 & 61.36 & 11.74 & 30.24\\
        & SAR~\cite{niutowards} & 82.08 & 71.31 & 2.76 & 13.72 & 85.36 & 76.00 & 2.09 & 7.69 & 68.42 & 53.23 & 12.7 & 33.81\\
        & GraTa~\cite{chen2025gradient} & 82.42 & 71.55 & 2.90 & 12.90 & 84.02 & 74.04 & 2.29 & 7.94 & 75.92 & 62.53 & 11.03 & 29.52\\
        & COME~\cite{zhangcome} & 84.00 & 73.85 & 2.64 & 12.77 & 86.35 & 77.38 & 2.01 & 7.39 & 75.52 & 62.36 & 10.95 & 29.15\\
        & SPA~\cite{niuself} & 83.94 & 73.77 & 2.69 & 12.81 & 83.60 & 73.70 & 2.26 & 8.19 & 63.34 & 48.85 & 13.53 & 34.04\\
        \hline
        \multirow{5}{*}{\makecell{ ATTA\\(10\%)}} & Random~\cite{settles2009active} & 83.25 & 72.98 & 2.64 & 12.89 & 86.40 & 77.46 & 1.93 & 7.15 & \cellcolor{yellow!20}78.01 & \cellcolor{yellow!20}65.77 & \cellcolor{yellow!20}9.99 & \cellcolor{yellow!20}26.29\\
        & Entropy~\cite{settles2009active} & \cellcolor{orange!20}85.94 & \cellcolor{orange!20}76.66 & \cellcolor{orange!20}2.30 & \cellcolor{orange!20}12.36 & \cellcolor{orange!20}88.45 & \cellcolor{orange!20}80.36 & \cellcolor{orange!20}1.66 & \cellcolor{orange!20}6.63 & \cellcolor{orange!20}78.90 & \cellcolor{orange!20}66.86 & \cellcolor{orange!20}9.08 & \cellcolor{orange!20}24.71\\
        & SimATTA~\cite{gui2024active} & 84.87 & 75.18 & 2.47 & \cellcolor{yellow!20}12.46 & 86.94 & 78.26 & 1.85 & 7.00 & 77.16 & 64.01 & 10.34 & 27.34\\
        & EATTA~\cite{wang2025effortless} & \cellcolor{yellow!20}85.22 & \cellcolor{yellow!20}75.64 & \cellcolor{yellow!20}2.35 & 12.58 & \cellcolor{yellow!20}88.31 & \cellcolor{yellow!20}80.13 & \cellcolor{yellow!20}1.68 & \cellcolor{yellow!20}6.74 &73.03 & 58.79 & 11.24 & 30.01\\
        & EviATTA (Ours) & \cellcolor{red!20}89.26 & \cellcolor{red!20}81.42  & \cellcolor{red!20}1.80 & \cellcolor{red!20}10.61 & \cellcolor{red!20}90.40 & \cellcolor{red!20}83.04 & \cellcolor{red!20}1.48 & \cellcolor{red!20}5.81 & \cellcolor{red!20}90.39 & \cellcolor{red!20}82.90 & \cellcolor{red!20}4.55 & \cellcolor{red!20}12.34\\
        \hline
        \multirow{5}{*}{\makecell{ATTA\\(20\%)}} & Random~\cite{settles2009active} & 83.80 & 73.62 & 2.59 & 12.73 & 86.40 & 77.46 & 1.93 & 7.15 & \cellcolor{orange!20}78.74 & \cellcolor{orange!20}66.50 & \cellcolor{yellow!20}9.78 & \cellcolor{yellow!20}25.88\\
        & Entropy~\cite{settles2009active} & \cellcolor{orange!20}86.70 & \cellcolor{orange!20}77.71 & \cellcolor{orange!20}2.18 & \cellcolor{orange!20}12.11 & \cellcolor{orange!20}88.96 & \cellcolor{orange!20}81.05 & \cellcolor{orange!20}1.59 & \cellcolor{orange!20}6.43 & \cellcolor{yellow!20}78.30 & \cellcolor{yellow!20}65.67 & \cellcolor{orange!20}9.34 & \cellcolor{orange!20}25.18\\
        & SimATTA~\cite{gui2024active} & 85.15 & 75.56 & 2.40 & \cellcolor{yellow!20}12.41 & 87.30 & 78.77 & 1.80 & 7.02 & 78.02 & 65.17 & 9.81 & 26.11\\
        & EATTA~\cite{wang2025effortless} & \cellcolor{yellow!20}85.47 & \cellcolor{yellow!20}76.00 & \cellcolor{yellow!20}2.36 & 12.42 & \cellcolor{yellow!20}88.60 & \cellcolor{yellow!20}80.50 & \cellcolor{yellow!20}1.65 & \cellcolor{yellow!20}6.63 & 74.96 & 60.90 & 10.69 & 28.87\\
        & EviATTA (Ours) & \cellcolor{red!20}89.61 & \cellcolor{red!20}81.99 & \cellcolor{red!20}1.69 & \cellcolor{red!20}10.30 & \cellcolor{red!20}90.81 & \cellcolor{red!20}83.67 & \cellcolor{red!20}1.41 & \cellcolor{red!20}5.64 & \cellcolor{red!20}90.97 & \cellcolor{red!20}83.80 & \cellcolor{red!20}4.28 & \cellcolor{red!20}11.69\\
        \hline
        \hline
        & \multicolumn{1}{c}{\multirow{2}{*}{Method}} & \multicolumn{4}{c}{REFUGE} & \multicolumn{4}{c}{RUNMC} & \multicolumn{4}{c}{PANTHER}\\
        \cmidrule(lr){3-6}\cmidrule(lr){7-10}\cmidrule(lr){11-14}
        & & Dice$\uparrow$ & Jaccard$\uparrow$ & ASD$\downarrow$ & HD95$\downarrow$ & Dice$\uparrow$ & Jaccard$\uparrow$ & ASD$\downarrow$ & HD95$\downarrow$ & Dice$\uparrow$ & Jaccard$\uparrow$ & ASD$\downarrow$ & HD95$\downarrow$\\
        \hline
        Zero-shot & SAM-Med2D & \cellcolor{gray!20}81.08 & \cellcolor{gray!20}68.92 & \cellcolor{gray!20}9.90 & \cellcolor{gray!20}28.61 & \cellcolor{gray!20}81.71 & \cellcolor{gray!20}71.65 & \cellcolor{gray!20}5.37 & \cellcolor{gray!20}16.39 & \cellcolor{gray!20}73.79 & \cellcolor{gray!20}59.84 & \cellcolor{gray!20}2.95 & \cellcolor{gray!20}9.20\\
        \hline
        \multirow{6}{*}{TTA} & TENT~\cite{wang2020tent} & 85.71 & 75.63 & 8.30 & 24.71 & 85.09 & 76.06 & 4.77 & 14.82 & 69.33 & 53.92 & 3.05 & 9.86\\
        & CoTTA~\cite{wang2022continual} & 79.39 & 66.68 & 9.93 & 29.00 & 79.95 & 69.25 & 5.68 & 17.32 & 75.78 & 62.23 & 2.84 & 8.77\\
        & SAR~\cite{niutowards} & 83.06 & 71.82 & 9.23 & 27.02 & 83.37 & 73.81 & 5.14 & 15.82 & 69.41 & 55.00 & 3.07 & 9.92\\
        & GraTa~\cite{chen2025gradient} & 87.29 & 77.85 & 8.06 & 24.61 & 84.04 & 74.76 & 4.95 & 15.31 & 75.25 & 61.56 & 2.96 & 9.11\\
        & COME~\cite{zhangcome} & 82.02 & 70.22 & 9.71 & 28.24 & 82.65 & 72.87 & 5.26 & 16.13 & 73.92 & 60.00 & 2.90 & 9.10\\
        & SPA~\cite{niuself} & 85.11 & 74.73 & 8.72 & 26.32 & 76.76 & 65.58 & 5.53 & 16.79 &72.03 & 58.48 & 11.75 & 30.56\\
        \hline
        \multirow{5}{*}{\makecell{ ATTA\\(10\%)}} & Random~\cite{settles2009active} & 86.97 & 77.54 & 7.64 & 23.35 & 84.67 & 75.62 & 4.86 & 14.92 & \cellcolor{orange!20}77.59 & \cellcolor{orange!20}64.45 & 2.82 & \cellcolor{orange!20}8.55\\
        & Entropy~\cite{settles2009active} & \cellcolor{yellow!20}88.71 & \cellcolor{yellow!20}80.24 & \cellcolor{yellow!20}6.37 & \cellcolor{yellow!20}19.37 & \cellcolor{orange!20}88.89 & \cellcolor{orange!20}81.22 & \cellcolor{orange!20}3.69 & \cellcolor{orange!20}11.35 & \cellcolor{yellow!20}76.73 & \cellcolor{yellow!20}63.26 & \cellcolor{orange!20}2.64 & 8.60\\
        & SimATTA~\cite{gui2024active} & 88.32 & 79.64 & 6.57 & 19.49 & 86.56 & 78.20 & 4.27 & 13.09 & 76.65 & 63.23 & \cellcolor{yellow!20}2.72 & \cellcolor{yellow!20}8.59\\
        & EATTA~\cite{wang2025effortless} & \cellcolor{orange!20}89.89 & \cellcolor{orange!20}82.12 & \cellcolor{orange!20}5.50 & \cellcolor{orange!20}16.06 & \cellcolor{yellow!20}87.59 & \cellcolor{yellow!20}79.54 & \cellcolor{yellow!20}4.02 & \cellcolor{yellow!20}12.25 & 75.01 & 61.16 & 2.74 & 9.07\\
        & EviATTA (Ours) & \cellcolor{red!20}91.25 & \cellcolor{red!20}84.37 & \cellcolor{red!20}4.33  & \cellcolor{red!20}11.68 & \cellcolor{red!20}90.15 & \cellcolor{red!20}82.82 & \cellcolor{red!20}3.36 & \cellcolor{red!20}9.50 & \cellcolor{red!20}80.14 & \cellcolor{red!20}67.72 & \cellcolor{red!20}2.49 & \cellcolor{red!20}7.74\\
        \hline
        \multirow{5}{*}{\makecell{ ATTA\\(20\%)}} & Random~\cite{settles2009active} & 86.67 & 77.11 & 7.66 & 23.69 & 87.16 & 78.95 & 4.28 & 13.07 & \cellcolor{yellow!20}77.92 & \cellcolor{yellow!20}64.85 & 2.80 & 8.56\\
        & Entropy~\cite{settles2009active} & \cellcolor{yellow!20}90.09 & \cellcolor{orange!20}82.45 & \cellcolor{yellow!20}5.23 & \cellcolor{yellow!20}15.35 & \cellcolor{orange!20}90.07 & \cellcolor{orange!20}82.99 & \cellcolor{orange!20}3.17 & \cellcolor{orange!20}9.82 & 77.91 & 64.78 & \cellcolor{orange!20}2.56 & \cellcolor{yellow!20}8.42\\
        & SimATTA~\cite{gui2024active} & 89.63 & 81.76 & 5.83 & 17.84 & 87.15 & 79.08 & 3.96 & 12.22 & \cellcolor{orange!20}78.25 & \cellcolor{orange!20}65.26 & \cellcolor{yellow!20}2.61 & \cellcolor{orange!20}8.35\\
        & EATTA~\cite{wang2025effortless} & \cellcolor{orange!20}90.11 & \cellcolor{yellow!20}82.41 & \cellcolor{orange!20}5.18 & \cellcolor{orange!20}14.34 & \cellcolor{yellow!20}88.85 & \cellcolor{yellow!20}81.27 & \cellcolor{yellow!20}3.65 & \cellcolor{yellow!20}11.29 & 75.88 & 62.26 & 2.63 & 8.90\\
        & EviATTA (Ours) & \cellcolor{red!20}91.70 & \cellcolor{red!20}85.11 & \cellcolor{red!20}3.99 & \cellcolor{red!20}10.55 & \cellcolor{red!20}90.77 & \cellcolor{red!20}83.75 & \cellcolor{red!20}3.09 & \cellcolor{red!20}8.82 & \cellcolor{red!20}81.13 & \cellcolor{red!20}69.04 & \cellcolor{red!20}2.42 & \cellcolor{red!20}7.53\\
        \toprule
        
    \end{tabularx}
    
    \label{tab:main_result}
\end{table*}

\begin{table*}[htbp]
    \centering
    \caption{Performance comparison with SOTA TTA and ATTA methods under instance-wise setting.}
    \setlength{\tabcolsep}{2mm}
    \begin{tabularx}{\linewidth}{clcccccccccccc}
        \bottomrule
         \multicolumn{1}{c}{\multirow{2}{*}{Type}} & \multicolumn{1}{c}{\multirow{2}{*}{Method}} & \multicolumn{4}{c}{BraTS-SSA} & \multicolumn{4}{c}{BraTS-PED} & \multicolumn{4}{c}{FHPsAOP}\\
        \cmidrule(lr){3-6}\cmidrule(lr){7-10}\cmidrule(lr){11-14}
        & & Dice$\uparrow$ & Jaccard$\uparrow$ & ASD$\downarrow$ & HD95$\downarrow$ & Dice$\uparrow$ & Jaccard$\uparrow$ & ASD$\downarrow$ & HD95$\downarrow$ & Dice$\uparrow$ & Jaccard$\uparrow$ & ASD$\downarrow$ & HD95$\downarrow$\\
        \hline
        Zero-shot & SAM-Med2D~\cite{cheng2023sammed2d} & \cellcolor{gray!20}82.34 & \cellcolor{gray!20}71.64 & \cellcolor{gray!20}2.73 & \cellcolor{gray!20}13.10 & \cellcolor{gray!20}84.84 & \cellcolor{gray!20}75.32 & \cellcolor{gray!20}2.12 & \cellcolor{gray!20}7.77 & \cellcolor{gray!20}73.87 & \cellcolor{gray!20}60.84 & \cellcolor{gray!20}11.16 & \cellcolor{gray!20}29.58\\
        
        \hline
        \multirow{6}{*}{TTA} & TENT~\cite{wang2020tent} & 80.59 & 69.42 & 2.84 & 14.19 & 86.03 & 77.04 & 1.98 & 7.64 & 65.48 & 49.98 & 13.07 & 35.33\\
        & CoTTA~\cite{wang2022continual} & 84.61 & 74.61 & 2.68 & 12.91 & 86.01 & 76.7 & 2.16 & 7.68 & 72.06 & 58.78 & 12.48 & 31.65\\
        & SAR~\cite{niutowards} & 82.46 & 71.75 & 2.76 & 13.21 & 85.44 & 76.07 & 2.11 & 7.73 & 65.80 & 50.12 & 13.43 & 35.46\\
        & GraTa~\cite{chen2025gradient} & 82.31 & 71.63 & 2.72 & 12.89 & 84.24 & 74.62 & 2.07 & 7.67 & 70.62 & 58.21 & 10.07 & 30.40\\
        & COME~\cite{zhangcome} & 84.57 & 74.64 & 2.62 & 12.60 & 86.71 & 77.87 & 1.99 & 7.29 & 75.88 & 62.65 & 10.90 & 29.11\\
        & SPA~\cite{niuself} & 84.19 & 74.11 & 2.69 & 12.70 & 83.48 & 73.54 & 2.25 & 8.23 & 72.03 & 58.48 & 11.75 & 30.56\\
        \hline
        \multirow{5}{*}{ATTA} & Random~\cite{settles2009active} & 85.96 & 76.51 & 2.35 & 12.19 & 88.71 & 80.50 & 1.70 & 6.57 & \cellcolor{orange!20}86.61 & \cellcolor{orange!20}76.94 & \cellcolor{orange!20}6.41 & \cellcolor{orange!20}17.57\\
        & Entropy~\cite{settles2009active} & 86.06 & 76.97 & 2.10 & 12.02 & 89.10 & 81.36 & \cellcolor{yellow!20}1.52 & 6.41 & \cellcolor{yellow!20}83.93 & \cellcolor{yellow!20}73.32 & \cellcolor{yellow!20}6.95 & 19.46\\
        & SimATTA~\cite{gui2024active} & \cellcolor{orange!20}87.06 & \cellcolor{orange!20}78.43 & \cellcolor{orange!20}1.94 & \cellcolor{orange!20}11.15 & \cellcolor{orange!20}89.76 & \cellcolor{orange!20}82.36 & \cellcolor{red!20}1.43 & \cellcolor{red!20}5.77 & 83.35 & 72.58 & 7.00 & \cellcolor{yellow!20}19.34\\
        & EATTA~\cite{wang2025effortless} & \cellcolor{yellow!20}86.62 & \cellcolor{yellow!20}77.75 & \cellcolor{yellow!20}2.04 & \cellcolor{yellow!20}12.00 & \cellcolor{yellow!20}89.20 & \cellcolor{yellow!20}81.54 & \cellcolor{orange!20}1.51 & \cellcolor{yellow!20}6.38 & 82.19 & 71.17 & 7.38 & 20.89\\
        & EviATTA (Ours) & \cellcolor{red!20}89.83 & \cellcolor{red!20}82.28 & \cellcolor{red!20}1.63 & \cellcolor{red!20}10.19 & \cellcolor{red!20}90.35 & \cellcolor{red!20}82.85 & 1.59 & \cellcolor{orange!20}6.06 & \cellcolor{red!20}91.43 & \cellcolor{red!20}84.34 & \cellcolor{red!20}4.32 & \cellcolor{red!20}12.07\\
        \hline
        \hline
        & \multicolumn{1}{c}{\multirow{2}{*}{Method}} & \multicolumn{4}{c}{REFUGE} & \multicolumn{4}{c}{RUNMC} & \multicolumn{4}{c}{PANTHER}\\
        \cmidrule(lr){3-6}\cmidrule(lr){7-10}\cmidrule(lr){11-14}
        & & Dice$\uparrow$ & Jaccard$\uparrow$ & ASD$\downarrow$ & HD95$\downarrow$ & Dice$\uparrow$ & Jaccard$\uparrow$ & ASD$\downarrow$ & HD95$\downarrow$ & Dice$\uparrow$ & Jaccard$\uparrow$ & ASD$\downarrow$ & HD95$\downarrow$\\
        \hline
        Zero-shot & SAM-Med2D~\cite{cheng2023sammed2d} & \cellcolor{gray!20}81.08 & \cellcolor{gray!20}68.92 & \cellcolor{gray!20}9.90 & \cellcolor{gray!20}28.61 & \cellcolor{gray!20}81.71 & \cellcolor{gray!20}71.65 & \cellcolor{gray!20}5.37 & \cellcolor{gray!20}16.39 & \cellcolor{gray!20}73.79 & \cellcolor{gray!20}59.84 & \cellcolor{gray!20}2.95 & \cellcolor{gray!20}9.20\\
        \hline
        \multirow{6}{*}{TTA} & TENT~\cite{wang2020tent} & 88.57 & 80.00 & 6.45 & 18.49 & 84.42 & 75.18 & 4.97 & 15.32 & 66.48 & 51.96 & 3.16 & 10.39\\
        & CoTTA~\cite{wang2022continual} & 78.99 & 66.15 & 9.94 & 28.98 & 80.82 & 70.39 & 5.55 & 17.00 & 75.62 & 62.04 & 2.88 & 8.89\\
        & SAR~\cite{niutowards} & 86.52 & 76.96 & 7.58 & 22.59 & 83.00 & 73.07 & 5.48 & 16.79 & 66.36 & 51.72 & 3.24 & 10.35\\
        & GraTa~\cite{chen2025gradient} & 87.19 & 78.01 & 7.13 & 20.69 & 81.92 & 73.41 & 3.93 & 11.97 & 72.74 & 58.79 & 2.92 & 9.53\\
        & COME~\cite{zhangcome} & 83.04 & 71.68 & 9.41 & 27.47 & 82.48 & 72.65 & 5.29 & 16.19 & 73.88 & 59.94 & 2.91 & 9.13\\
        & SPA~\cite{niuself} & 85.89 & 75.90 & 8.33 & 25.29 & 78.25 & 67.37 & 5.52 & 16.90 & 72.60 & 58.45 & 2.98 & 9.39\\
        \hline
        \multirow{5}{*}{ATTA} & Random~\cite{settles2009active} & 87.92 & 79.11 & 5.94 & 17.78 & \cellcolor{orange!20}90.09 & \cellcolor{yellow!20}83.10 & 3.17 & 9.84 & 79.29 & 66.57 & 2.90 & 8.88\\
        & Entropy~\cite{settles2009active} & \cellcolor{yellow!20}91.78 & \cellcolor{yellow!20}85.26 & \cellcolor{yellow!20}3.73 & \cellcolor{yellow!20}10.29 & \cellcolor{yellow!20}89.40 & 82.65 & \cellcolor{yellow!20}2.81 & \cellcolor{yellow!20}8.76 & \cellcolor{orange!20}80.62 & \cellcolor{orange!20}68.42 & 2.58 & 8.26\\
        & SimATTA~\cite{gui2024active} & \cellcolor{orange!20}92.59 & \cellcolor{orange!20}86.46 & \cellcolor{orange!20}3.28 & \cellcolor{orange!20}8.13 & 89.92 & \cellcolor{orange!20}83.43 & \cellcolor{orange!20}2.71 & \cellcolor{orange!20}8.59 & \cellcolor{yellow!20}80.25 & \cellcolor{yellow!20}67.91 & \cellcolor{orange!20}2.49 & \cellcolor{orange!20}7.98\\
        & EATTA~\cite{wang2025effortless} & 90.93 & 83.91 & 4.48 & 13.91 & 87.24 & 79.92 & 3.18 & 10.12 & 79.62 & 67.13 & \cellcolor{red!20}2.46 & \cellcolor{yellow!20}8.40\\
        & EviATTA (Ours) & \cellcolor{red!20}94.62 & \cellcolor{red!20}89.88 & \cellcolor{red!20}2.26 & \cellcolor{red!20}5.45 & \cellcolor{red!20}93.16 & \cellcolor{red!20}87.66 & \cellcolor{red!20}2.17 & \cellcolor{red!20}6.26 & \cellcolor{red!20}80.97 & \cellcolor{red!20}68.83 & \cellcolor{yellow!20}2.53 & \cellcolor{red!20}7.89\\
        \toprule
        
    \end{tabularx}
    
    \label{tab:main_result2}
\end{table*}

\subsection{Datasets and Evaluation Metrics}
We evaluated EviATTA on six medical image segmentation datasets covering multiple modalities, organs, and lesions: BraTS-SSA~\cite{adewole2023brain}, BraTS-PED~\cite{adewole2023brain}, FHPsAOP~\cite{kucs2024medsegbench}, REFUGE~\cite{orlando2020refuge}, RUNMC~\cite{bloch2015nciisbi}, and PANTHER~\cite{betancourt2025panther}. 
\textbf{BraTS-SSA}~\cite{adewole2023brain} contains 60 adult glioma mpMRI scans from Sub-Saharan Africa, which yield 3,695 axial T2-FLAIR slices for whole-tumor segmentation. 
\textbf{BraTS-PED}~\cite{adewole2023brain} includes 99 mp-MRI scans of pediatric tumors, providing 3,739 T2-FLAIR slices for whole-tumor segmentation.
\textbf{FHPsAOP}~\cite{kucs2024medsegbench} contains 4,000 ultrasound images for symphysis-fetal head analysis. We focus on fetal head segmentation in this study.
\textbf{REFUGE}~\cite{orlando2020refuge} consists of 800 retinal fundus images for optic disc and cup segmentation. We focus on optic disc segmentation in this study.
\textbf{RUNMC}~\cite{bloch2015nciisbi} is a prostate segmentation dataset with 30 T2-weighted MRI scans, from which 420 axial slices are extracted for evaluation.
\textbf{PANTHER}~\cite{betancourt2025panther} includes 50 T2-weighted MR-Linac scans for pancreas and tumor segmentation. Both structures are treated as foreground, and 1,906 axial slices are used for experiments.

The segmentation results are evaluated using four metrics, including Dice Similarity Coefficient (Dice, \%), Jaccard Index (Jaccard, \%), Average Surface Distance (ASD, in voxels), and 95\% Hausdorff Distance (HD95, in voxels). Higher Dice and Jaccard scores indicate superior segmentation performance, whereas lower values are desirable for ASD and HD95.

\subsection{Implementation Details}
In this study, we adopt SAM-Med2D~\cite{cheng2023sammed2d} as the backbone due to its strong zero-shot performance and public availability. All images are resized to $256\times 256$ pixels. Following~\cite{cheng2023sammed2d}, we use the bounding box derived from the ground-truth as the initial prompt. We evaluate EviATTA under two adaptation regimes: (1) batch-wise adaptation with $|\mathcal{B}|{=}32$, and (2) instance-wise adaptation with $|\mathcal{B}|{=}1$. For both settings, we perform one-step optimization per test batch following~\cite{chen2025gradient}. We use Adam with a learning rate of $1e{-4}$ for batch-wise adaptation and $1e{-}5$ for instance-wise adaptation. LoRA modules with rank $r=4$ are inserted into the image encoder~\cite{chen2025personalized}. For batch-wise adaptation, we evaluate EviATTA under two annotation budgets. Specifically, 10\% and 20\% of samples in each test batch are selected, with $M=5$ annotated pixels per selected sample. For instance-wise adaptation, we query the annotation of $M=5$ points per test sample. We maintain a replay buffer of size 128, from which we randomly replay 16 samples per optimization step in batch-wise adaptation and a single sample in the instance-wise setting. All experiments are conducted on a single NVIDIA L40S GPU.

\subsection{Comparison Methods}
We compared EviATTA with eleven state-of-the-art (SOTA) methods, including a zero-shot baseline SAM-Med2D~\cite{cheng2023sammed2d}, six test-time adaptation methods, and four active test-time adaptation methods. The \textbf{test-time adaptation} methods comprise test-time entropy minimization (TENT)~\cite{wang2020tent}, continual test-time domain adaptation (CoTTA)~\cite{wang2022continual}, sharpness-aware and reliable entropy minimization (SAR)~\cite{niutowards}, gradient alignment-based test-time adaptation (GraTa)~\cite{chen2025gradient}, conservatively minimizing entropy (COME)~\cite{zhangcome}, and self-bootstrapping TTA (SPA)~\cite{niuself}. The \textbf{active test-time adaptation} methods cover random sampling (Random)~\cite{budd2021survey}, entropy sampling (Entropy)~\cite{budd2021survey}, simple yet effective ATTA (SimATTA)~\cite{gui2024active}, and effortless active test-time adaptation (EATTA)~\cite{wang2025effortless}. Owing to the absence of BN layers, ODES~\cite{islam2025odes} cannot be applied to SAM for BN-based sample selection and is therefore excluded from the main experiments.

\subsection{Results under Batch-wise Adaptation}
The results under batch-wise adaptation are reported in Table~\ref{tab:main_result}, where \textit{ATTA (10\%)} and \textit{ATTA (20\%)} denote selecting 10\% and 20\% of samples per batch for sparse annotation, respectively. The top three results are highlighted as \colorbox{red!20}{first}, \colorbox{orange!20}{second}, and \colorbox{yellow!20}{third}. Overall, TTA methods exhibit unstable performance under distribution shifts and can even cause negative adaptation. In contrast, ATTA methods consistently outperform both zero-shot and TTA baselines by introducing sparse yet reliable supervision. As the annotation budget increases from 10\% to 20\%, most ATTA methods further improve, and EviATTA remains the best-performing method on all six datasets across Dice, Jaccard, ASD, and HD95.

On \textbf{BraTS-SSA}, TTA methods yield only marginal improvements over the zero-shot baseline. The best TTA method (CoTTA) improves Dice by 1.75\% and Jaccard by 2.26\%, with negligible reductions in ASD and HD95. In contrast, EviATTA improves Dice by 6.92\% and Jaccard by 9.78\% over SAM-Med2D and surpasses the second-best ATTA method (Entropy) by 3.32\% in Dice and 4.76\% in Jaccard under \textit{ATTA (10\%)}. On \textbf{BraTS-PED}, which shows substantial shifts from adult BraTS data, EviATTA also achieves the best results, with Dice and Jaccard gains of 5.56\% and 7.72\% over the zero-shot baseline. On \textbf{FHPsAOP}, where TTA methods frequently degrade performance, existing ATTA baselines recover only moderate gains of approximately 5\% in Dice and 6\% in Jaccard, whereas EviATTA achieves substantially larger improvements of 16.5\% and 22.06\%, respectively. EviATTA also consistently outperforms all baselines on \textbf{REFUGE}, \textbf{RUNMC}, and \textbf{PANTHER}, achieving Dice improvements of 10.17\%, 8.44\%, and 6.35\% under \textit{ATTA (10\%)}. Notably, increasing the annotation budget from \textit{ATTA (10\%)} to \textit{ATTA (20\%)} yields a further average Dice improvement of 0.57\% for EviATTA across all six benchmarks, demonstrating its effectiveness in leveraging additional sparse annotations.

\subsection{Results under Instance-wise Adaptation}
Table~\ref{tab:main_result2} reports instance-wise adaptation results, where ATTA methods are restricted to pixel-wise selection. In this setting, TTA baselines suffer from significant instability, frequently underperforming even the zero-shot baseline. This indicates that unreliable self-supervision poses a risk of error accumulation, ultimately leading to negative adaptation. In contrast, EviATTA demonstrates superior performance in both volumetric overlap and boundary accuracy. Compared to the zero-shot baseline, EviATTA achieves an average Dice improvement of 10.46\% and an average HD95 reduction of 9.56 voxels across six benchmarks. Furthermore, it consistently outperforms the second-best ATTA method, yielding an additional average Dice gain of 2.27\% and an average HD95 reduction of 1.88 voxels across all datasets. These substantial performance gains can be attributed to the proposed pixel selection strategy and dual consistency regularization, which empower EviATTA to annotate informative pixels and fully exploit sparse annotations.

\subsection{Ablation Study}
Table~\ref{tab:component} presents a component-wise ablation on BraTS-SSA. Without adaptation, the zero-shot SAM-Med2D yields 82.34\% in Dice and 13.10 voxels in HD95. Solely treating sparse annotations as spatial prompts provides marginal overlap gains and fails to improve boundary quality, which indicates that sparse supervision alone is insufficient. In contrast, introducing progressive prompt consistency $\mathcal{L}_{prompt}$ markedly boosts Dice to 88.02\% and reduces HD95 to 11.35 voxels. The addition of the replay scheme $\mathcal{L}_{replay}$ further enhances Dice by 0.27\% and drops HD95 by 0.22 voxels, which confirms its efficacy in maximizing sparse annotation utility. Finally, incorporating variational feature consistency $\mathcal{L}_{var}$ achieves the best performance, reaching 89.26\% in Dice and 10.61 voxels in HD95. Overall, the full EviATTA framework delivers a gain of 6.92\% in Dice and a reduction of 2.49 voxels in HD95 compared to the zero-shot baseline. These results validate the complementary effect of all proposed components.
\begin{table}[htbp]
    \centering
    \caption{Effect of each component.}
    \setlength{\tabcolsep}{1mm}
    \begin{tabular}{cccccccc}
    \bottomrule
        \multirow{2}{*}{Annotation}& \multicolumn{3}{c}{Consistency Regularization}& \multicolumn{4}{c}{BraTS-SSA}\\
        & $\mathcal{L}_{prompt}$ & $\mathcal{L}_{replay}$ & $\mathcal{L}_{var}$ & Dice$\uparrow$ & Jaccard$\uparrow$ & ASD$\downarrow$ & HD95$\downarrow$\\
        \hline
        & & & & 82.34 & 71.64 & 2.73 & 13.10 \\
        \checkmark & & & & 83.25 & 72.83 & 2.67 & 13.15\\
        \checkmark & \checkmark & & & 88.02 & 79.46 & 2.10 & 11.35\\
        \checkmark & \checkmark & \checkmark & & 88.29 & 79.95 & 2.00 & 11.13\\
        \rowcolor{gray!20}\checkmark & \checkmark & \checkmark & \checkmark & 89.26 & 81.42 & 1.80 & 10.61\\
    \toprule
    \end{tabular}
    \label{tab:component}
\end{table}

\section{Discussion}
The following experiments adopt a batch-wise setup, with 10\% of the most informative samples sparsely annotated.

\subsection{Effect of Sampling Strategy}
\begin{figure}[htbp]
    \centering
    \includegraphics[width=\linewidth]{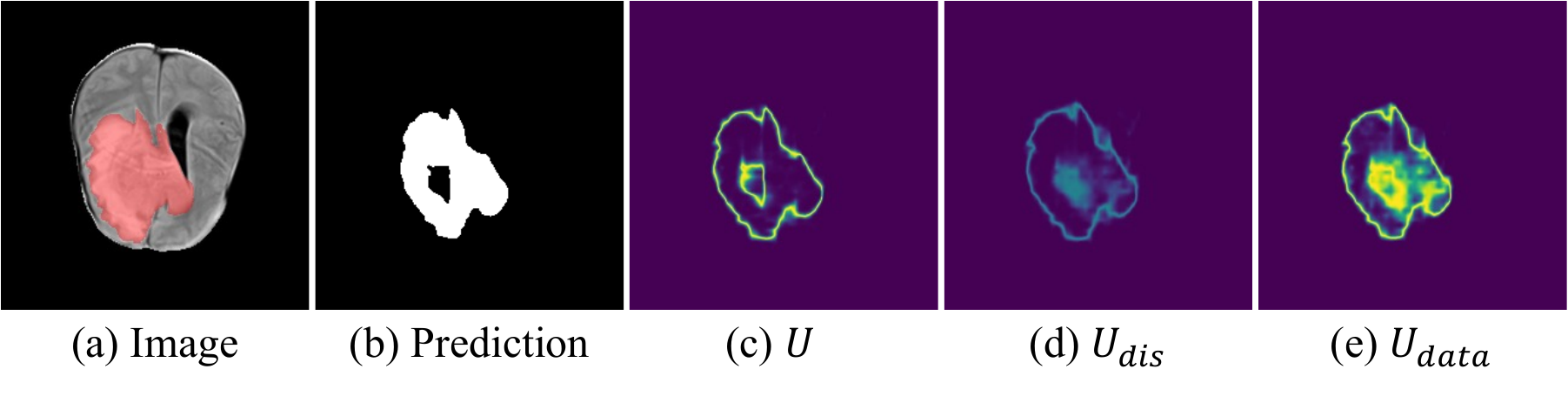}
    \caption{Visualization of uncertainty map. $U$ denotes predictive uncertainty. $U_{dis}$ and $U_{data}$ represent distribution and data uncertainties.}
    \label{fig:visualization_u}
\end{figure}

Table~\ref{tab:uncertainty} analyzes the impact of different sampling strategies on BraTS-SSA. Compared to the zero-shot baseline, random sampling at both sample-level and pixel-level yields only marginal gains, improving Dice by 0.91\%. Replacing random scores with overall predictive uncertainty ($U/U$) boosts Dice by 2.69\% and reduces HD95 by 0.53 voxels, underscoring the efficacy of uncertainty-guided annotation. We then investigate decoupling uncertainty into distribution ($U_{dis}$) and data ($U_{data}$) components. Applying either decomposed uncertainty across both levels surpasses the overall uncertainty, yielding an approximate 1.44\% rise in Dice and a 0.7-voxel reduction in HD95. The optimal performance is achieved by the asymmetric configuration ($U_{dis}/U_{data}$), which reaches 89.26\% in Dice and 10.61 voxels in HD95. This configuration delivers a gain of 3.32\% in Dice and a reduction of 1.75 voxels in HD95 compared to the overall uncertainty setting, which also outperforms the swapped variant $U_{data}/U_{dis}$. These findings confirm that sample selection should use $U_{dis}$ to capture distribution shifts, whereas pixel annotation should rely on $U_{data}$ to resolve inherent data ambiguities. Finally, we provide qualitative support in Fig.~\ref{fig:visualization_u} to highlight that decomposed uncertainties focus on different regions, while $U_{data}$ more precisely localizes erroneous regions for sparse annotation.
\begin{table}[htbp]
    \centering
    \caption{Effect of Sampling Strategy.}
    
    \begin{tabular}{cccccccc}
    \bottomrule
        \multicolumn{2}{c}{Uncertainty Estimation} & \multicolumn{4}{c}{BraTS-SSA}\\
        Sample-wise & Pixel-wise & Dice$\uparrow$ & Jaccard$\uparrow$ & ASD$\downarrow$ & HD95$\downarrow$\\
        \hline
        - & - & 82.34 & 71.64 & 2.73 & 13.10\\
        Random & Random & 83.25 & 72.98 & 2.64 & 12.89\\
        $U$ & $U$ & 85.94 & 76.66 & 2.30 & 12.36\\
        $U_{data}$ & $U_{data}$ & 87.34 & 78.36 & 2.35 & 11.71\\
        $U_{dis}$ & $U_{dis}$ & 87.38 & 78.39 & 2.33 & 11.66\\
        $U_{data}$ & $U_{dis}$ & 89.06 & 81.09 & 1.85 & 10.71\\
        \rowcolor{gray!20}$U_{dis}$ & $U_{data}$ & 89.26 & 81.42 & 1.80 & 10.61\\
    \toprule
    \end{tabular}
    
    \label{tab:uncertainty}
\end{table}

\subsection{Effect of Pixel Annotation Budget}
\begin{figure}[htbp]
    \centering
    \includegraphics[width=\linewidth]{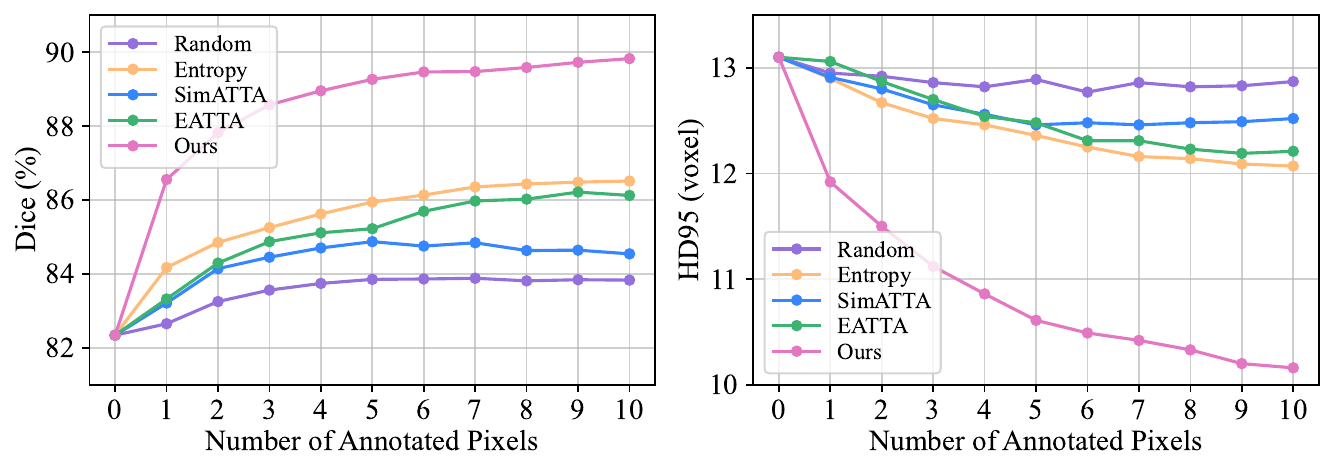}
    \caption{Effect of pixel annotation budgets on BraTS-SSA.}
    \label{fig:point_num}
\end{figure}

We explore the effect of pixel annotation budget on BraTS-SSA. As reported in Fig.~\ref{fig:point_num}, the segmentation performance generally improves with more annotated pixels, since more pixel annotations serve as point prompts to provide stronger spatial supervision. Compared with existing ATTA methods, EviATTA consistently yields higher Dice and Jaccard and lower ASD and HD95 under different annotation budgets. The results indicate that EviATTA identifies more informative samples and pixels and extensively exploits the limited sparse annotations. To balance segmentation performance and annotation costs, we set the annotation budget to five points per selected sample in this study.

\subsection{Effect of Replay Buffer}
\begin{figure}[htbp]
    \centering
    \includegraphics[width=\linewidth]{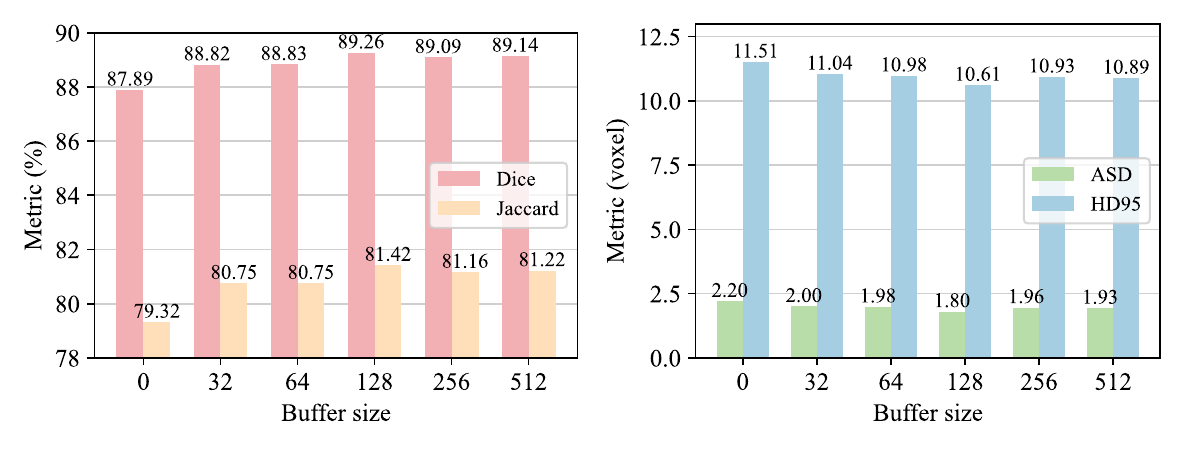}
    \caption{Effect of buffer size on BraTS-SSA.}
    \label{fig:replay}
\end{figure}

Fig.~\ref{fig:replay} shows the effect of the replay buffer on BraTS-SSA. Without replay, EviATTA achieves a Dice of 87.89\% and HD95 of 11.51 voxels. Introducing a buffer of 32 samples improves Dice by 0.93\% and reduces HD95 by 0.47 voxels. The performance remains stable across different buffer sizes and peaks when the buffer size is set to 128. Therefore, we adopt a buffer size of 128 in the experiments. These results demonstrate that the replay buffer effectively leverages the limited sparse annotations.

\subsection{Effect of LoRA Placement}
Table~\ref{tab:lora_pos} compares different LoRA placements on the query ($\boldsymbol{W}_Q$), key ($\boldsymbol{W}_K$), value ($\boldsymbol{W}_V$), and output ($\boldsymbol{W}_O$) projections. Without LoRA, EviATTA achieves 83.25\% in Dice, 72.83\% in Jaccard, 2.67 voxels in ASD, and 13.15 voxels in HD95. Introducing LoRA into any individual projection consistently improves performance, confirming the efficacy of parameter-efficient adaptation at test time. Specifically, applying LoRA to $\boldsymbol{W}_V$ or $\boldsymbol{W}_O$ proves more effective than tuning $\boldsymbol{W}_Q$ or $\boldsymbol{W}_K$, improving Dice by ${\sim}2\%$ and reducing HD95 by ${\sim}1$ voxel. Among all configurations, applying LoRA to $\boldsymbol{W}_O$ achieves the highest volumetric overlap and boundary precision, improving Dice by 6.01\% and reducing HD95 by 2.54 voxels compared with the baseline. In contrast, simultaneously adapting multiple projections does not yield cumulative gains. $\boldsymbol{W}_Q{+}\boldsymbol{W}_K$ shows no improvement over single-projection settings, while $\boldsymbol{W}_Q{+}\boldsymbol{W}_V$ performs slightly worse than using $\boldsymbol{W}_V$ alone. Therefore, we adopt LoRA on the output projection in all experiments.
\begin{table}[htbp]
    \centering
    \caption{Effect of LoRA Placement}
    \begin{tabular}{cccccccc}
    \bottomrule
        \multicolumn{4}{c}{Weight Matrix} & \multicolumn{4}{c}{BraTS-SSA}\\
        $\boldsymbol{W}_Q$ & $\boldsymbol{W}_K$ & $\boldsymbol{W}_V$ & $\boldsymbol{W}_O$ & Dice$\uparrow$ & Jaccard$\uparrow$ & ASD$\downarrow$ & HD95$\downarrow$\\
        \hline
        &&&&83.25 & 72.83 & 2.67 & 13.15\\
        \checkmark & & & & 87.23 & 78.43 & 2.27 & 11.60\\
        & \checkmark & &  & 87.23 & 78.44 & 2.27 & 11.60\\
        & & \checkmark & & 89.09 & 81.13 & 1.86 & 10.69\\
        \rowcolor{gray!20}& & & \checkmark & 89.26 & 81.42 & 1.80 & 10.61\\
        \checkmark & \checkmark & & & 87.22 & 78.40 & 2.27 & 11.62\\
        \checkmark & & \checkmark & & 88.90 & 80.93 & 1.94 & 10.77\\
    \toprule
    \end{tabular}
    
    \label{tab:lora_pos}
\end{table}

\begin{figure*}[t]
    \centering
    \includegraphics[width=\linewidth]{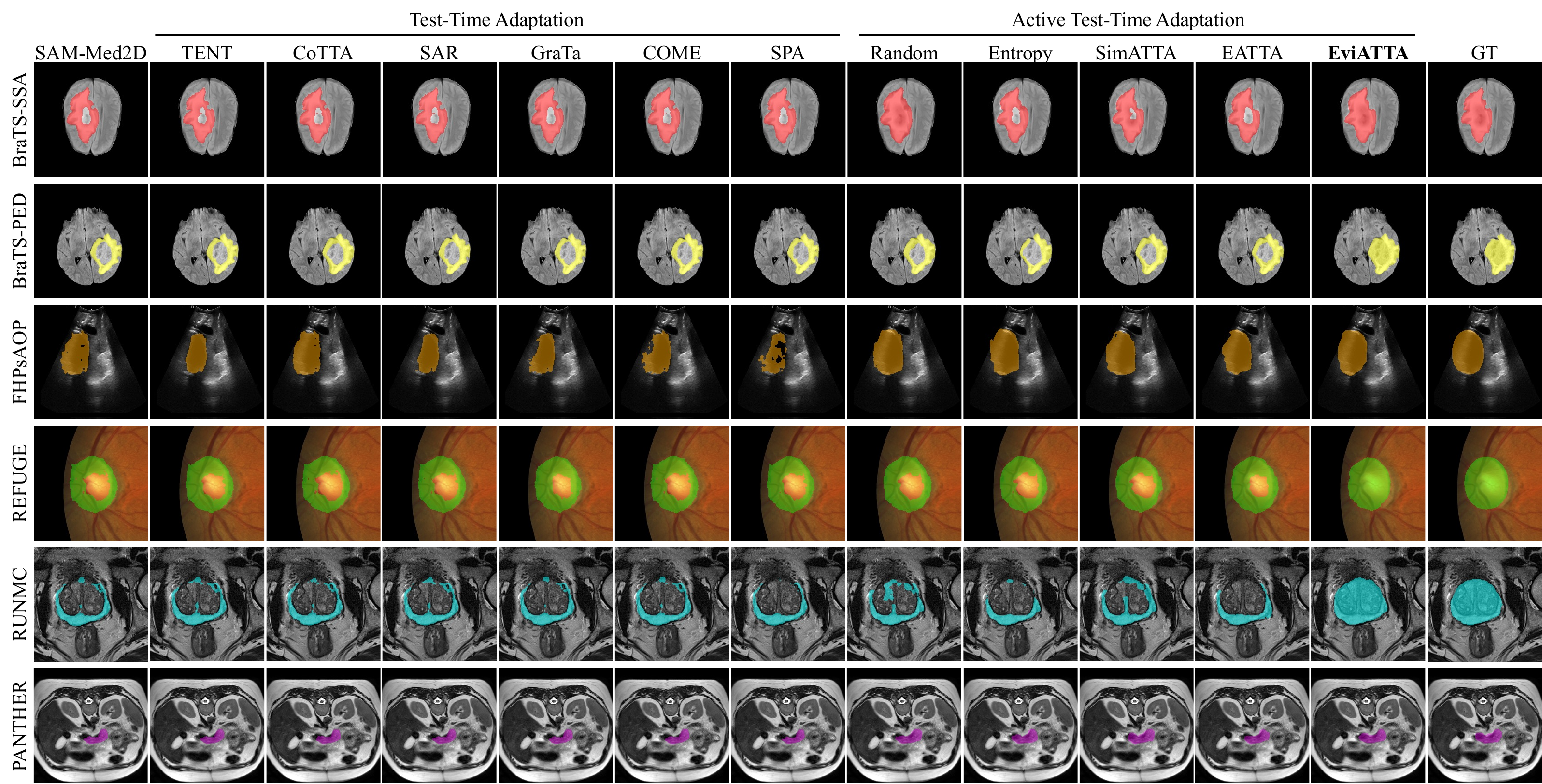}
    \caption{Qualitative segmentation results of different methods across six medical image segmentation datasets.}
    \label{fig:visualization}
\end{figure*}

\subsection{Effect of Initial Prompt}
\begin{figure}[h]
    \centering
    \includegraphics[width=\linewidth]{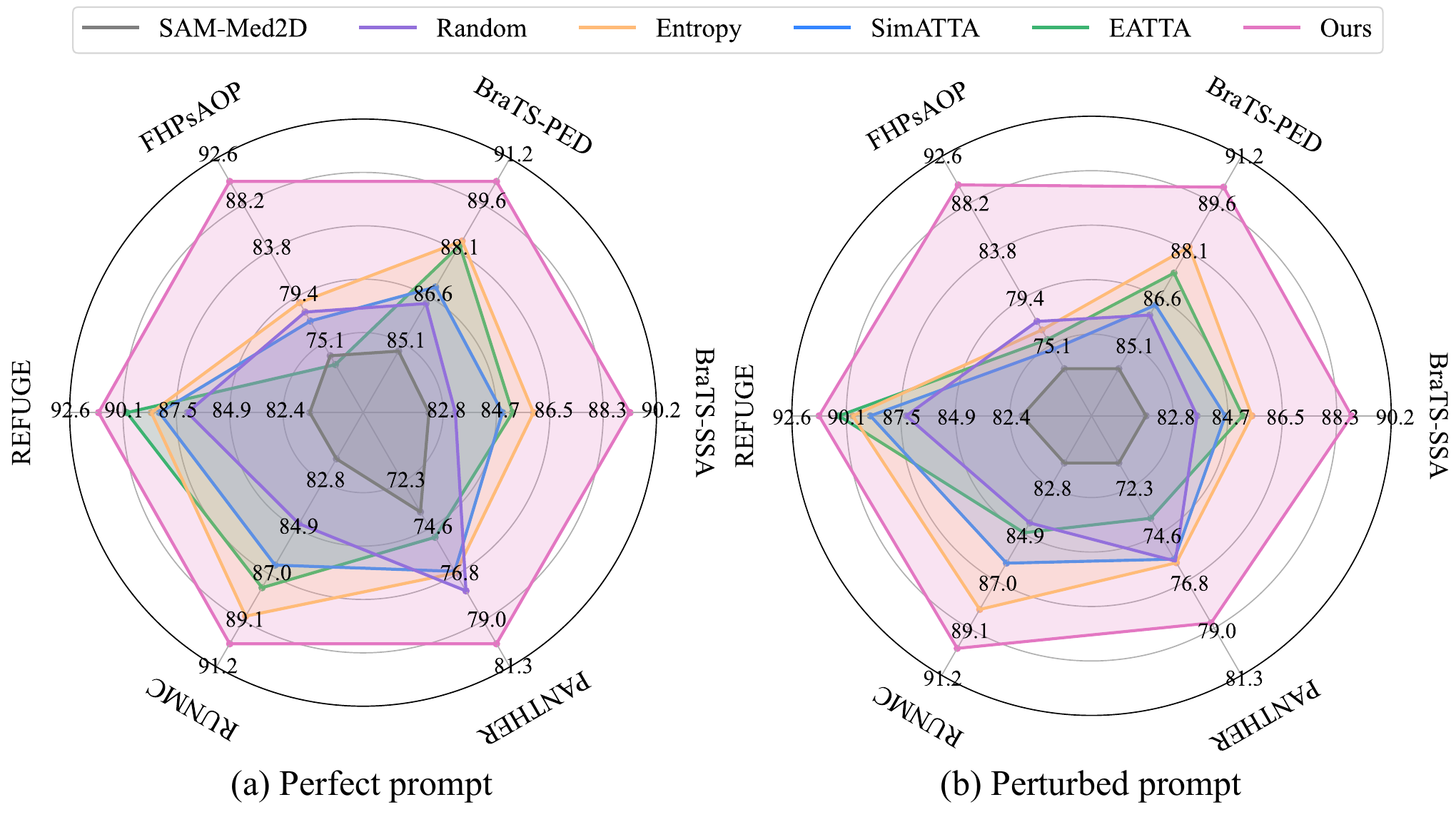}
    \caption{Effect of initial prompt on six datasets.}
    \label{fig:prompt}
\end{figure}

As shown in Fig.~\ref{fig:prompt}, we evaluate the effect of initial prompt quality on adaptation across six benchmarks. We compare two prompt settings: a perfect prompt and a perturbed prompt, where the perturbed prompt is generated by adding random noise to the coordinates of the perfect bounding box. Prompt perturbation causes clear performance degradation for zero-shot SAM-Med2D. In contrast, ATTA methods are more stable because sparse expert annotations provide corrective supervision during adaptation. Across all datasets and both prompt settings, EviATTA consistently achieves the best performance, demonstrating robustness against varying prompt qualities.

\subsection{Visualization of Point Prompts}
\begin{figure}[h]
    \centering
    \includegraphics[width=\linewidth]{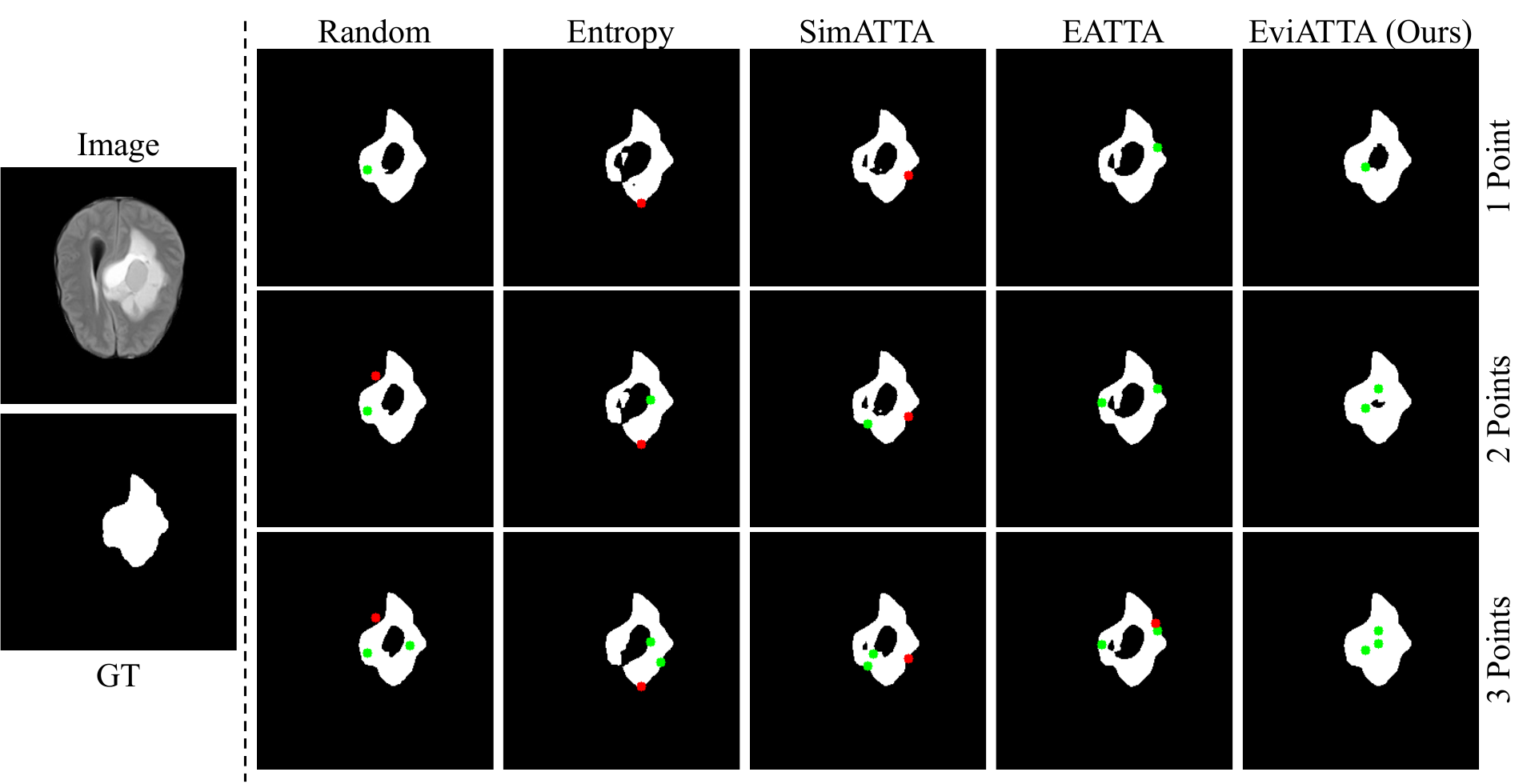}
    \caption{Segmentation results under different numbers of point prompts.}
    \label{fig:point_prompt}
\end{figure}

Fig.~\ref{fig:point_prompt} provides a qualitative comparison under different numbers of point prompts, where green and red points denote positive and negative prompts, respectively. Conventional ATTA methods tend to query low-impact pixels, leading to limited refinement. In contrast, EviATTA selects more informative and structurally critical points, enabling more effective error correction with minimal annotations. Under the same annotation budget, EviATTA produces the most accurate masks, which qualitatively validates its reliable uncertainty estimation and effective utilization of sparse annotations.

\subsection{Visualization of Segmentation Results}
Fig.~\ref{fig:visualization} presents qualitative comparisons of segmentation results across six datasets. The zero-shot SAM-Med2D often produces incomplete structures and boundary errors, which are only partially addressed by unsupervised TTA methods. While ATTA baselines improve visual quality with sparse annotations, their refinements remain inconsistent across different cases. In contrast, EviATTA consistently generates more complete structures and cleaner boundaries with fewer errors across diverse anatomical structures and imaging modalities. These visual results align well with our quantitative findings, confirming that EviATTA improves adaptation reliability with minimal annotation effort.

\section{Conclusion}
We have proposed EviATTA, the first ATTA method for medical SAMs to alleviate error accumulation with minimal annotation cost. First, we introduced Dirichlet-based evidential modeling to decompose overall uncertainty into distribution and data uncertainties. Then, we designed a hierarchical evidential sampling strategy, which considers both sample-wise distribution uncertainty and pixel-wise data uncertainty to annotate the most informative pixels. We further proposed a dual consistency regularization to exploit sparse annotations and maintain adaptation stability. Extensive experiments on six benchmarks have demonstrated the superiority of EviATTA in improving adaptation reliability with minimal expert feedback. In future work, we will extend EviATTA to 3D volumetric and video data to build a unified ATTA framework for medical foundation segmentation models.

\bibliographystyle{IEEEtran}
\bibliography{tmi}

\begin{thebibliography}{10}
\providecommand{\url}[1]{#1}
\csname url@samestyle\endcsname
\providecommand{\newblock}{\relax}
\providecommand{\bibinfo}[2]{#2}
\providecommand{\BIBentrySTDinterwordspacing}{\spaceskip=0pt\relax}
\providecommand{\BIBentryALTinterwordstretchfactor}{4}
\providecommand{\BIBentryALTinterwordspacing}{\spaceskip=\fontdimen2\font plus
\BIBentryALTinterwordstretchfactor\fontdimen3\font minus \fontdimen4\font\relax}
\providecommand{\BIBforeignlanguage}[2]{{%
\expandafter\ifx\csname l@#1\endcsname\relax
\typeout{** WARNING: IEEEtran.bst: No hyphenation pattern has been}%
\typeout{** loaded for the language `#1'. Using the pattern for}%
\typeout{** the default language instead.}%
\else
\language=\csname l@#1\endcsname
\fi
#2}}
\providecommand{\BIBdecl}{\relax}
\BIBdecl

\bibitem{cheng2023sammed2d}
J.~Cheng \emph{et~al.}, ``{SAM-Med2D},'' \emph{arXiv preprint arXiv:2308.16184}, 2023.

\bibitem{ma2024segment}
J.~Ma, Y.~He, F.~Li, L.~Han, C.~You, and B.~Wang, ``Segment anything in medical images,'' \emph{Nature Communications}, vol.~15, no.~1, p. 654, 2024.

\bibitem{wu2025medical}
J.~Wu \emph{et~al.}, ``Medical {SAM} adapter: Adapting segment anything model for medical image segmentation,'' \emph{Medical image analysis}, vol. 102, p. 103547, 2025.

\bibitem{zhu2024medical}
J.~Zhu, A.~Hamdi, Y.~Qi, Y.~Jin, and J.~Wu, ``Medical {SAM} 2: Segment medical images as video via segment anything model 2,'' \emph{arXiv preprint arXiv:2408.00874}, 2024.

\bibitem{huang2024segment}
Y.~Huang \emph{et~al.}, ``Segment anything model for medical images?'' \emph{Medical Image Analysis}, vol.~92, p. 103061, 2024.

\bibitem{ma2025tsar}
B.~Ma \emph{et~al.}, ``{TSAR}: A two-stage approach to motion artifact reduction in octa images,'' \emph{Pattern Recognition}, p. 112364, 2025.

\bibitem{xia2024enhancing}
Y.~Xia, B.~Ma, Q.~Dou, and Y.~Xia, ``Enhancing federated learning performance fairness via collaboration graph-based reinforcement learning,'' in \emph{MICCAI}.\hskip 1em plus 0.5em minus 0.4em\relax Springer, 2024, pp. 263--272.

\bibitem{chen2025test}
K.~Chen \emph{et~al.}, ``Test-time adaptation for foundation medical segmentation model without parametric updates,'' in \emph{ICCV}, 2025, pp. 20\,075--20\,084.

\bibitem{fu2024cosam}
Y.~Fu, Z.~Chen, Y.~Ye, X.~Lei, Z.~Wang, and Y.~Xia, ``{CoSAM}: self-correcting sam for domain generalization in 2d medical image segmentation,'' \emph{arXiv preprint arXiv:2411.10136}, 2024.

\bibitem{ma2023federated}
B.~Ma, Y.~Feng, G.~Chen, C.~Li, and Y.~Xia, ``Federated adaptive reweighting for medical image classification,'' \emph{Pattern Recognition}, vol. 144, p. 109880, 2023.

\bibitem{huang2025fly}
T.~Huang, T.~Zhou, W.~Xie, S.~Wang, Q.~Dou, and Y.~Zhang, ``On-the-fly improving segment anything for medical image segmentation using auxiliary online learning,'' \emph{IEEE Transactions on Medical Imaging}, 2025.

\bibitem{liang2025comprehensive}
J.~Liang, R.~He, and T.~Tan, ``A comprehensive survey on test-time adaptation under distribution shifts,'' \emph{International Journal of Computer Vision}, vol. 133, no.~1, pp. 31--64, 2025.

\bibitem{wang2020tent}
D.~Wang, E.~Shelhamer, S.~Liu, B.~Olshausen, and T.~Darrell, ``{TENT}: Fully test-time adaptation by entropy minimization,'' in \emph{ICLR}, 2021.

\bibitem{zhangcome}
Q.~Zhang, Y.~Bian, X.~Kong, P.~Zhao, and C.~Zhang, ``{COME}: Test-time adaption by conservatively minimizing entropy,'' in \emph{ICLR}, 2024.

\bibitem{chen2025gradient}
Z.~Chen, Y.~Ye, Y.~Pan, and Y.~Xia, ``Gradient alignment improves test-time adaptation for medical image segmentation,'' in \emph{AAAI}, vol.~39, no.~3, 2025, pp. 2429--2437.

\bibitem{wang2022continual}
Q.~Wang, O.~Fink, L.~Van~Gool, and D.~Dai, ``Continual test-time domain adaptation,'' in \emph{CVPR}, 2022, pp. 7201--7211.

\bibitem{niuself}
S.~Niu, G.~Chen, P.~Zhao, T.~Wang, P.~Wu, and Z.~Shen, ``Self-bootstrapping for versatile test-time adaptation,'' in \emph{ICML}, 2025.

\bibitem{wu2025sam}
J.~Wu \emph{et~al.}, ``{SAM}-aware test-time adaptation for universal medical image segmentation,'' \emph{arXiv preprint arXiv:2506.05221}, 2025.

\bibitem{wang2025effortless}
G.~Wang and C.~Ding, ``Effortless active labeling for long-term test-time adaptation,'' in \emph{CVPR}, 2025, pp. 25\,633--25\,642.

\bibitem{niutowards}
S.~Niu \emph{et~al.}, ``Towards stable test-time adaptation in dynamic wild world,'' in \emph{ICLR}, 2023.

\bibitem{chen2022contrastive}
D.~Chen, D.~Wang, T.~Darrell, and S.~Ebrahimi, ``Contrastive test-time adaptation,'' in \emph{CVPR}, 2022, pp. 295--305.

\bibitem{huselective}
Y.~Hu, C.~Qiao, X.~Geng, and N.~Xu, ``Selective label enhancement learning for test-time adaptation,'' in \emph{ICLR}, 2025.

\bibitem{niu2022efficient}
S.~Niu \emph{et~al.}, ``Efficient test-time model adaptation without forgetting,'' in \emph{ICML}, 2022, pp. 16\,888--16\,905.

\bibitem{gui2024active}
S.~Gui, X.~Li, and S.~Ji, ``Active test-time adaptation: Theoretical analyses and an algorithm,'' in \emph{ICLR}, 2024.

\bibitem{wang2024comprehensive}
H.~Wang, Q.~Jin, S.~Li, S.~Liu, M.~Wang, and Z.~Song, ``A comprehensive survey on deep active learning in medical image analysis,'' \emph{Medical Image Analysis}, vol.~95, p. 103201, 2024.

\bibitem{liexploring}
Y.~Li, Y.~Su, X.~Yang, K.~Jia, and X.~Xu, ``Exploring human-in-the-loop test-time adaptation by synergizing active learning and model selection,'' \emph{Transactions on Machine Learning Research}, 2025.

\bibitem{leetest}
T.~Lee, S.~Chottananurak, J.~Kim, J.~Shin, T.~Gong, and S.-J. Lee, ``Test-time adaptation with binary feedback,'' in \emph{ICML}, 2025.

\bibitem{tanexposing}
J.~Tan, F.~Lyu, C.~Ni, W.~Feng, F.~Hu, and R.~Yao, ``Exposing mixture and annotating confusion for active universal test-time adaptation,'' in \emph{ICLR}, 2026.

\bibitem{shi2025annotation}
T.~Shi, F.~Lyu, and S.~Peng, ``Annotation-efficient active test-time adaptation with conformal prediction,'' \emph{arXiv preprint arXiv:2509.25692}, 2025.

\bibitem{yuan2023few}
L.~Yuan, S.~Li, Z.~He, and B.~Xie, ``Few clicks suffice: Active test-time adaptation for semantic segmentation,'' \emph{arXiv preprint arXiv:2312.01835}, 2023.

\bibitem{islam2025odes}
M.~S. Islam \emph{et~al.}, ``{ODES}: Online domain adaptation with expert guidance for medical image segmentation,'' in \emph{MICCAI}.\hskip 1em plus 0.5em minus 0.4em\relax Springer, 2025, pp. 359--370.

\bibitem{watson2023explaining}
D.~Watson, J.~O'Hara, N.~Tax, R.~Mudd, and I.~Guy, ``Explaining predictive uncertainty with information theoretic shapley values,'' in \emph{NeurIPS}, vol.~36, 2023, pp. 7330--7350.

\bibitem{chen2024think}
J.~Chen, B.~Ma, H.~Cui, and Y.~Xia, ``Think twice before selection: Federated evidential active learning for medical image analysis with domain shifts,'' in \emph{CVPR}, 2024, pp. 11\,439--11\,449.

\bibitem{sensoy2018evidential}
M.~Sensoy, L.~Kaplan, and M.~Kandemir, ``Evidential deep learning to quantify classification uncertainty,'' in \emph{NeurIPS}, vol.~31, 2018.

\bibitem{chen2024fedevi}
J.~Chen, B.~Ma, H.~Cui, and Y.~Xia, ``{FedEvi}: Improving federated medical image segmentation via evidential weight aggregation,'' in \emph{MICCAI}.\hskip 1em plus 0.5em minus 0.4em\relax Springer, 2024, pp. 361--372.

\bibitem{xiedirichlet}
M.~Xie, S.~Li, R.~Zhang, and C.~H. Liu, ``Dirichlet-based uncertainty calibration for active domain adaptation,'' in \emph{ICLR}, 2023.

\bibitem{hu2024towards}
S.~Hu, Z.~Liao, Z.~Liu, and Y.~Xia, ``Towards clinician-preferred segmentation: Leveraging human-in-the-loop for test time adaptation in medical image segmentation,'' \emph{arXiv preprint arXiv:2405.08270}, 2024.

\bibitem{chen2025active}
J.~Chen, B.~Ma, H.~Cui, J.~Zhang, and Y.~Xia, ``Active learning based on temporal difference of gradient flow in thoracic disease diagnosis,'' \emph{IEEE Journal of Biomedical and Health Informatics}, 2025.

\bibitem{josang2016subjective}
A.~J{\o}sang, \emph{Subjective logic}.\hskip 1em plus 0.5em minus 0.4em\relax Springer, 2016, vol.~3.

\bibitem{malinin2018predictive}
A.~Malinin and M.~Gales, ``Predictive uncertainty estimation via prior networks,'' in \emph{NeurIPS}, vol.~31, 2018.

\bibitem{otsu1979threshold}
N.~Otsu \emph{et~al.}, ``A threshold selection method from gray-level histograms,'' \emph{Automatica}, vol.~11, no. 285-296, 1979.

\bibitem{ma2024vnas}
B.~Ma, J.~Zhang, Y.~Xia, and D.~Tao, ``{VNAS}: variational neural architecture search,'' \emph{International Journal of Computer Vision}, vol. 132, no.~9, pp. 3689--3713, 2024.

\bibitem{settles2009active}
B.~Settles, ``Active learning literature survey,'' 2009.

\bibitem{adewole2023brain}
M.~Adewole \emph{et~al.}, ``The brain tumor segmentation ({BraTS}) challenge 2023: Glioma segmentation in sub-saharan africa patient population ({BraTS-Africa}),'' \emph{ArXiv}, pp. arXiv--2305, 2023.

\bibitem{kucs2024medsegbench}
Z.~Ku{\c{s}} and M.~Aydin, ``{MedSegBench}: A comprehensive benchmark for medical image segmentation in diverse data modalities,'' \emph{Scientific Data}, vol.~11, no.~1, p. 1283, 2024.

\bibitem{orlando2020refuge}
J.~I. Orlando \emph{et~al.}, ``{REFUGE} challenge: A unified framework for evaluating automated methods for glaucoma assessment from fundus photographs,'' \emph{Medical Image Analysis}, vol.~59, p. 101570, 2020.

\bibitem{bloch2015nciisbi}
\BIBentryALTinterwordspacing
N.~Bloch \emph{et~al.}, ``{NCI-ISBI} 2013 challenge: Automated segmentation of prostate structures,'' The Cancer Imaging Archive, 2015. [Online]. Available: \url{http://doi.org/10.7937/K9/TCIA.2015.zF0vlOPv}
\BIBentrySTDinterwordspacing

\bibitem{betancourt2025panther}
\BIBentryALTinterwordspacing
A.~S. Betancourt~Tarifa, F.~Mahmood, U.~Bernchou, and P.~J. Koopmans, ``{PANTHER} challenge: Public training dataset,'' Zenodo, Apr. 2025. [Online]. Available: \url{https://doi.org/10.5281/zenodo.15192302}
\BIBentrySTDinterwordspacing

\bibitem{chen2025personalized}
J.~Chen, B.~Ma, Y.~Pan, B.~Pu, H.~Cui, and Y.~Xia, ``Personalized federated side-tuning for medical image classification,'' in \emph{MICCAI}.\hskip 1em plus 0.5em minus 0.4em\relax Springer, 2025, pp. 452--462.

\bibitem{budd2021survey}
S.~Budd, E.~C. Robinson, and B.~Kainz, ``A survey on active learning and human-in-the-loop deep learning for medical image analysis,'' \emph{Medical Image Analysis}, vol.~71, p. 102062, 2021.

\end{thebibliography}
\end{document}